\documentclass{article}

\usepackage{arxiv}
\pdfoutput=1 
\usepackage[utf8]{inputenc} 
\usepackage[T1]{fontenc}    
\usepackage{hyperref}       
\usepackage{url}            
\usepackage{booktabs}       
\usepackage{amsfonts}       
\usepackage{nicefrac}       
\usepackage{enumitem}
\usepackage{microtype}      
\usepackage{lipsum}
\usepackage{graphicx}
\usepackage{xcolor}
\usepackage{longtable}
\title{Swarm Learning: A Survey of Concepts, Applications, and Trends}

\author{
Elham Shammar \\
School of Cyber Science and Engineering, \\
Wuhan University, Wuhan, China\\
  \texttt{eshammar@whu.edu.cn} \\
\And 
Xiaohui Cui \\
School of Cyber Science and Engineering, \\
Wuhan University, Wuhan, China\\
  \texttt{xcui@whu.edu.cn} \\
\And
 Mohammed A. A. Al-qaness \\
  College of Physics and Electronic \\ Information Engineering\\
  Zhejiang Normal University\\
  Jinhua 321004, China\\
  \texttt{alqaness@zjnu.edu.cn} \\
}

\begin{document}
\maketitle

\begin{abstract}
Deep learning (DL) models have raised privacy and security concerns due to their reliance on large datasets on central servers. As the number of devices on the Internet of Things (IoT) increases, artificial intelligence (AI) will be crucial for resource management, data processing, and knowledge acquisition. To address those issues, federated learning (FL) has introduced a novel approach to building a versatile, large-scale machine learning (ML) framework that operates in a decentralized and hardware-agnostic manner. However, FL faces network bandwidth limitations and data breaches. To reduce the central dependency in FL and increase scalability, swarm learning (SL) has been proposed in collaboration with Hewlett Packard Enterprise (HPE). SL represents a decentralized ML framework that leverages blockchain technology for secure, scalable, and private data management. A blockchain-based network enables the exchange and aggregation of model parameters among participants, thus mitigating the risk of a single point of failure and eliminating communication bottlenecks. This survey introduces the principles of SL, its architectural design, and its fields of application. In addition, it highlights numerous research avenues that require further exploration by academic and industry communities to unlock the full potential and applications of SL.

\end{abstract}

 

 
\section{Introduction}

The number of Internet of Things (IoT) devices is expected to grow rapidly over the next five years. In 2019, a third of all IoT devices were used in healthcare, and this is predicted to increase to 40\%, representing \$6.2 trillion of the global IoT market by 2025 \cite{alsubaei2019iomt}. By 2030, there will likely be 29 billion IoT devices in different industries, and IoMT devices could potentially save \$300 billion, particularly in chronic disease management and telemedicine \cite{baho2023analysis, ghubaish2020recent}. This market is expected to generate \$135 billion in revenue by 2025. In addition, the global healthcare market is projected to reach \$6.2 trillion by 2028 \cite{dovgal2021swarm}, highlighting the need for advances in AI, resource management, data processing, and knowledge extraction, all driven by the rapid growth of 5G and Multi-Access Edge Computing (MEC) \cite{sun2021decentralized}.

Deep learning (DL) increases the power of IoT by allowing devices to handle large amounts of data, improve decision making through better pattern recognition, and enhance system performance in real-time applications such as smart cities, healthcare, and self-driving cars. However, these modern DL models raise serious concerns about privacy and security, mainly because they rely on centralized servers to store large datasets \cite{chen2023backdoor}.Although cloud-based local learning has certain advantages, there are drawbacks as well, such as increased security threats, traffic, and data redundancy. Data ownership and privacy concerns are another problem with traditional centralized learning approaches \cite{dovgal2021swarm}. One such solution is federated learning (FL), which allows decentralized cooperation while maintaining data privacy. However, FL has to contend with issues like bandwidth constraints and cyberthreats. Distributed FL (DFL) \cite{zhang2024decentralized} and swarm learning (SL), a new strategy created in collaboration with Hewlett Packard Enterprise (HPE) \cite{chen2023backdoor}, are two promising answers to these issues.

DFL and SL both aim to improve privacy and reduce reliance on centralized data storage. DFL allows model training on multiple nodes without needing a central server, making the process more flexible and decentralized \cite{beltran2023decentralized, sengupta2022blockchain, ghanem2022flobc}. On the other hand, SL takes advantage of blockchain to build a decentralized peer-to-peer network, which improves both security and privacy. By combining blockchain with federated learning, SL removes the central server altogether, allowing training to happen directly at each node, where only the parameter weights are shared. This approach reduces the risks associated with storing data in one central location. Blockchain's consensus mechanisms ensure the process remains secure, while SL uses smart contracts for more secure, privacy-friendly data sharing. Unlike traditional federated learning, which aggregates local gradients, SL uses a more secure and streamlined process that includes steps such as node registration, authentication, training, gradient sharing, and result aggregation using the Federated Average method \cite{madni2023blockchain, ARIKKAT2025107562, Wang2024}.

SL’s decentralized approach boosts fault tolerance, scalability, and privacy, making it a great fit for a range of sectors such as healthcare, automotive, finance, smart cities, edge computing, IoT, and the metaverse. In healthcare, SL ensures secure collaborative training while keeping patient data protected \cite{warnat2021swarm}. In industries, SL helps make faster and more informed decisions by allowing independent learning agents to work together, which is in good agreement with the goals of Industry 4.0 \cite{pongfai2020novel, pongfai2020pid, sun2023data, sun2023improved}. In the finance sector, SL plays a key role in detecting fraud while ensuring that customer information remains private \cite{john2023swarm}. For smart cities, it can improve things like traffic flow and public transport management. Furthermore, SL supports data ownership, helps meet regulatory standards, sparks innovation, and gives businesses a competitive edge by accelerating the speed with which they can bring new products to market \cite{yin2023multi}.

This survey offers a detailed analysis of SL, delving into its architecture, components, and practical applications. It highlights how SL’s integration with blockchain improves both security and scalability, while also examining its adaptability in dynamic real-world settings. Unlike the study by Evangelia et al. \cite{fragkou2024joint}, which does not provide much insight into practical deployment, this article offers a clearer picture of how scalable SL is and how it performs in real-world applications. Although Tajabadi et al. \cite{TAJABADI20243281} focus primarily on healthcare uses of SL, this work broadens the discussion by exploring how SL handles non-IID data, enhances security through blockchain and improves node coordination efficiency. In doing so, it fills the gaps left by earlier surveys, offering a more well-rounded view of SL's potential in decentralized learning systems.

This paper is motivated by the growing need to explore and unlock the full potential of SL as a decentralized, privacy-focused machine learning (ML) framework. We look closer at how SL can be implemented in real-world scenarios, assess its strengths and challenges, and examine how it is being applied across different industries. With data privacy becoming more critical and the growth of IoT devices, this survey aims to provide useful insights for researchers and professionals looking to leverage SL for secure, scalable, and efficient ML solutions in various domains.

\subsection{Paper objectives and contribution}
The key advantages of SL call for a deeper look at its current research and practical applications, as well as a careful identification of areas that need further development. This survey explores the potential of SL in decentralized learning environments, highlights both technical and operational challenges, and suggests possible directions for future advancements. Specifically, it focuses on how advanced cryptographic methods can enhance security and how SL can be adapted to new technologies such as edge computing and the Internet of Things.

This work consolidates existing knowledge, highlights research gaps, and outlines strategic directions for the broader adoption of SL. As a resource, this survey aims to guide scholars, researchers, and practitioners in expanding SL's application across various industries, thus maximizing its impact and utility.
\\
In summary, the main contribution of this paper can be presented as follows: 
\begin{itemize}
    \item We present the first survey paper in the field of SL. To the best of our knowledge, this paper is the first review on SL.
    \item We provide a comprehensive overview of the existing literature on SL and its current applications to give the readers a complete picture of this new and promising research direction. 
    \item We studied and analyzed the current applications of SL. We categorized them into healthcare, transportation, industry, robotic systems, smart homes, financial services, multimedia IoT, fake news detection, and Metaverse.  
    \item We present an in-depth analysis of the current limitations and challenges facing SL. We explore how these issues impact their development and deployment. In addition, we discuss potential future directions to improve SL technologies and applications. We suggest avenues for advancement and areas that are ready for further research to enhance the effectiveness and applicability of SL technologies.
\end{itemize}
\color{black}

\begin{figure}[htbp]
\centering
\includegraphics[width=.8\linewidth]{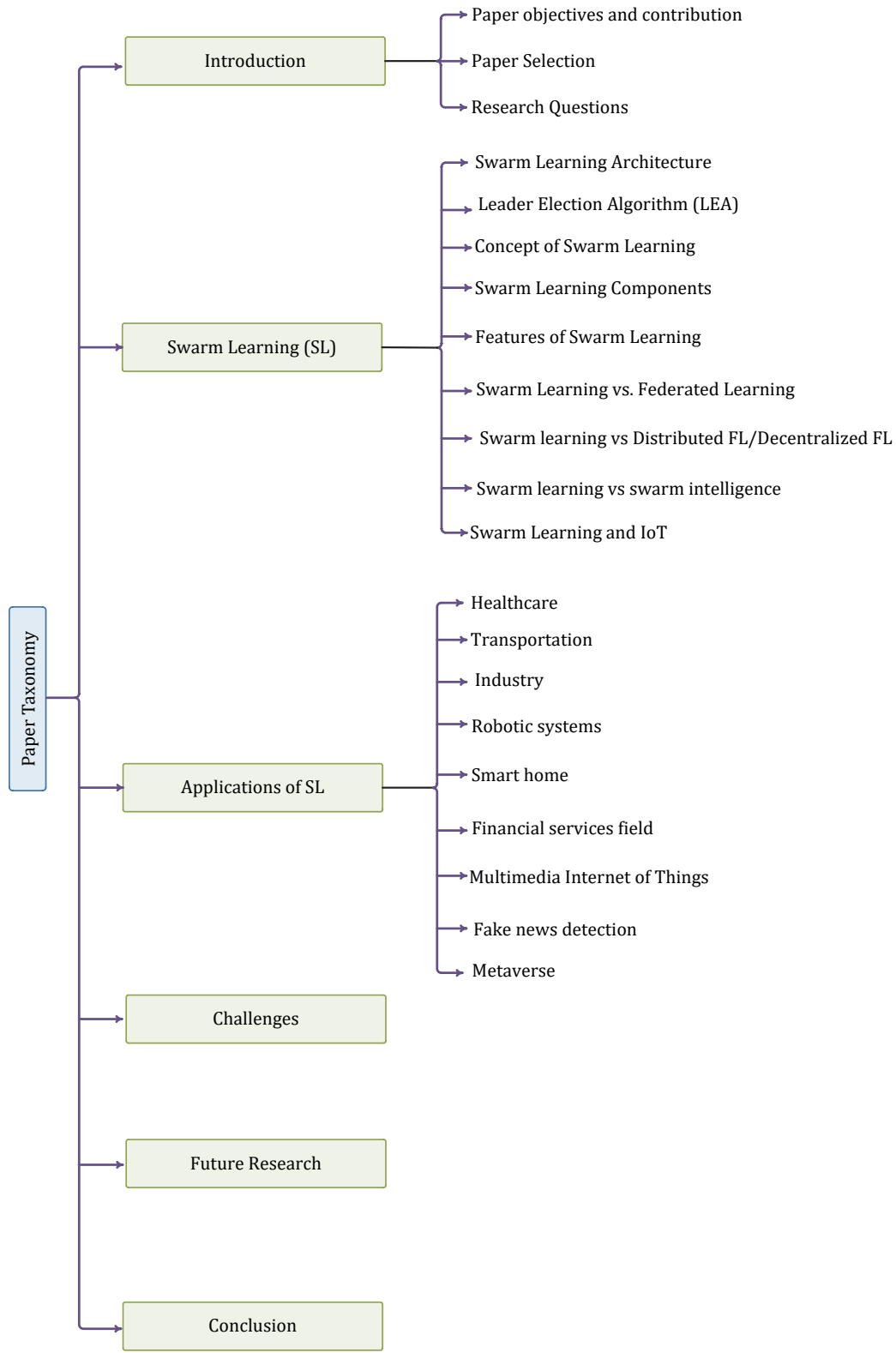} 
\caption{Paper Structure}
\label{fig:Taxonomy} 
\end{figure}

The structure of this paper, illustrated in Fig.\ref{fig:Taxonomy}, is as follows: Section 2 introduces SL, covering its fundamental concepts and components. Sections 3 and 4 discuss the applications of SL and the challenges associated with its implementation, respectively. Section 5 outlines potential future research directions for SL. Finally, the paper concludes in Section 6.

\section{Research Methodology}
The primary motivation for this study is to answer the research questions in Subsection 2.1. We aim to explore the core concepts, architecture, and components of SL, compare it with other decentralized learning approaches, and identify its real-world applications and challenges. 

\subsection{Research Questions}
The following research questions guide the exploration of Swarm Learning (SL), its architecture, applications, and challenges, aiming to deepen the understanding of this emerging decentralized learning approach.
\begin{enumerate}
    \item How do the key concepts, architecture, and components of SL function together in a distributed learning environment?
    \item How does SL compare with FL, distributed FL, decentralized FL, and swarm intelligence regarding performance, privacy, and scalability?
    \item What are the specific applications of SL in industries such as healthcare, finance, and IoT, and how do these applications benefit from its use?
    \item What are the primary challenges faced during the adoption and implementation of SL in real-world applications, particularly regarding data privacy, communication overhead, and system scalability?
\end{enumerate}

\subsection{Papers Selection}
We conducted a thorough search in six databases: IEEE, PubMed, Science Direct, Scopus, Springer, and Web of Science. We retrieved 30 papers from IEEE, 12 from PubMed, 129 from Science Direct, 87 from Scopus, 28 from Springer, and 56 from Web of Science. After screening for relevance to SL and excluding articles on swarm intelligence and optimization, we identified 84 papers that met our inclusion criteria.

\begin{figure}[htbp]
\centering
\includegraphics[width=0.8\linewidth]{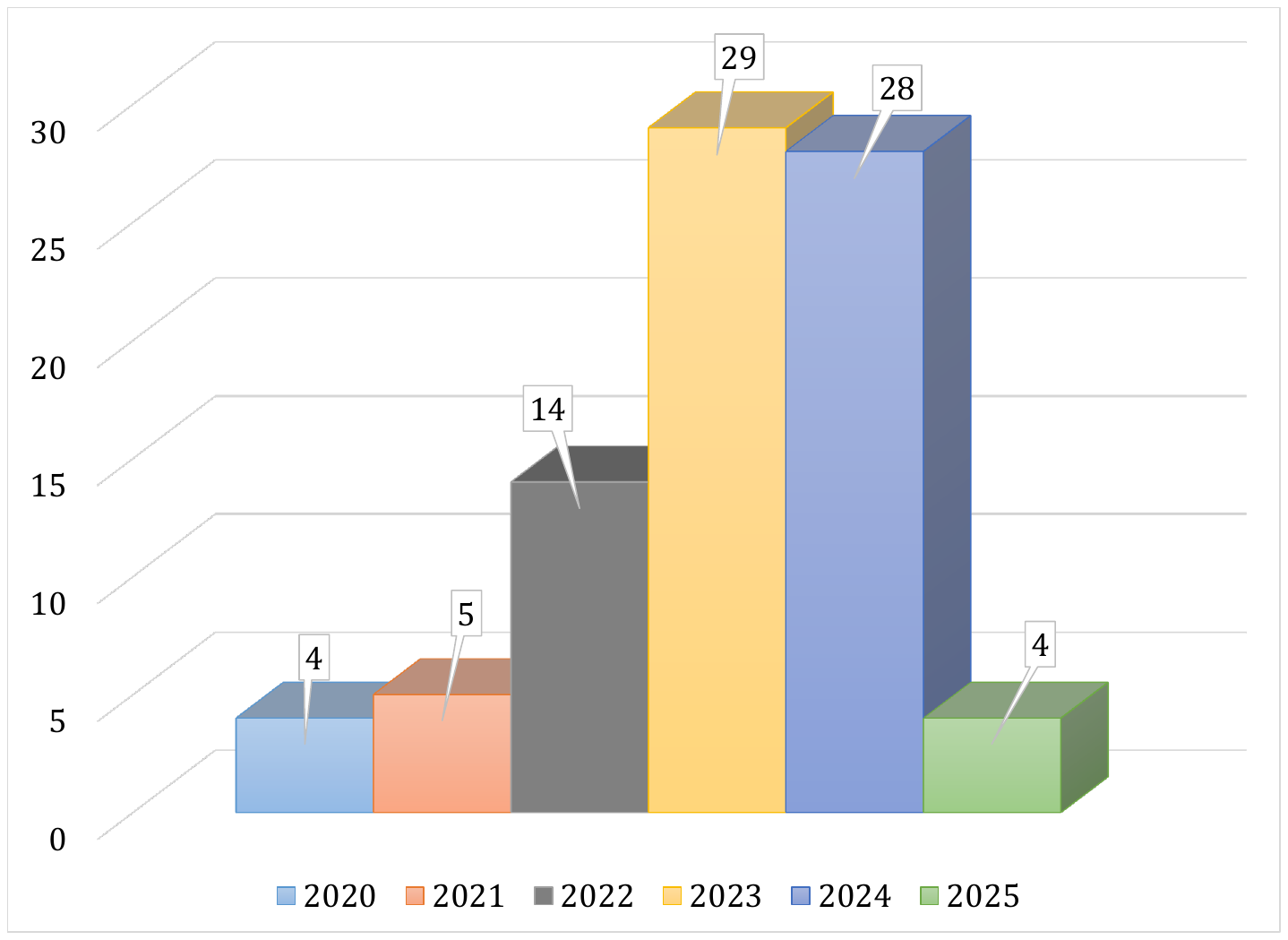} 
\caption{Annual increase in the number of Swarm Learning
research papers.}
\label{fig:Swarm learning Annual Increase} 
\end{figure}

The number of SL research papers has grown steadily, as shown in Fig.~\ref{fig:Swarm learning Annual Increase}. Starting with four papers in 2020, it increased to five in 2021, 14 in 2022, and a substantial increase to 29 in 2023. In 2024, there were 28 publications, reflecting the growing importance of SL. The number for 2025 stands at only four due to the survey being finalized in February 2025, providing a snapshot of SL research up to that time. This upward trend highlights the increasing academic interest in SL, driven by advances in computational power, data availability, the rise of IoT devices, privacy-preserving AI techniques, and the need for decentralized solutions in sectors such as healthcare, autonomous driving, and smart cities.

\section{Swarm Learning (SL)}
SL, developed in collaboration with HPE \cite{chen2023backdoor}, is a decentralized machine learning framework that enables the training of the model on the device without the need to transfer raw data. By keeping data localized on the data owner's site, SL substantially reduces data traffic by avoiding the transmission of raw data \cite{chen2021privacy}. Using blockchain technology, SL enhances privacy and security by exchanging only the model parameters and weights, not the actual data itself. This approach utilizes smart contracts to manage the training and updating of decentralized ML models using local user data, separating it from traditional centralized systems or even FL frameworks that rely on a central server to gather model updates \cite{john2023swarm}. Furthermore, SL incorporates advanced data privacy and security mechanisms, making it an ideal, flexible, and secure solution for content caching in modern network architectures \cite{yang2023swarm}.

SL employs a permissioned blockchain network and a decentralized hardware infrastructure to facilitate secure member onboarding, dynamic leader election, and efficient merging of model parameters. The system utilizes standardized AI engines within a distributed ML context to ensure secure and reliable operations. An SL library supports an iterative AI learning process that leverages decentralized data, adhering rigorously to the prevailing privacy and security standards \cite{fan2021fairness}. This structured approach secures data and streamlines the computational process between various network nodes.
 
Han et al. \cite{han2022demystifying} bridge the gap between SL theory and practice by conducting experiments on three public datasets, demonstrating the suitability of SL in various scenarios, even with unbalanced, polluted, or biased data. However, challenges like backdoor attacks, blockchain integration complexity, and computational overhead persist. Han et al. \cite{10701316} evaluated SL's performance in noisy and unbalanced datasets, focusing on scalability and resource utilization. Their work highlights SL's potential in decentralized environments while addressing real-world issues such as dataset distribution and deployment strategies.

Mantovani et al. \cite{10628751} introduced a Key Performance Indicator (KPI) to evaluate SL performance in distributed healthcare analysis. Their approach eliminates central server management by distributing model parameters across nodes and addressing privacy concerns. Using data from the Intensive Care Unit (ICU) of the Medical Information Mart for Intensive Care (MIMIC) Electronic Health Records (EHR) database, they demonstrated SL's potential to improve data privacy and collaboration in decentralized healthcare analysis.

SL addresses several key issues in FL and provides many benefits in security, privacy, and scalability. SL can envision ways to develop more secure, private, and faster distributed ML applications from different domains. To address gradient leakage and data privacy concerns in Federated Learning (FL), Madni et al. \cite{madni2023blockchain} developed a decentralized secure framework that combined blockchain with Swarm Learning (SL). SL ensures data privacy and model parameter secrecy, authenticating only trusted nodes and maintaining data integrity through blockchain mechanisms. Research shows that SL outperforms current anomaly detection methods in ML, offering better precision and mitigating gradient leakage, a key limitation in FL.

In two articles \cite{fan2023efficient}, \cite{fan2023robust}, Xu et al. tackled data heterogeneity, security, and communication bottlenecks in FL by proposing a robust edge learning framework for IoT devices. They introduced Byzantine-Robust and Communication-Efficient Distributed SL (CB-DSL), the first comprehensive theoretical analysis of FL with Particle Swarm Optimization (PSO). This framework includes a closed-form formula for assessing the convergence rate of CB-DSL, demonstrating its superiority over traditional FL methods like Federated Averaging (FedAvg) and provides a model divergence analysis for improving learning outcomes in non-IID scenarios.

The following subsections discuss the architecture, components, features, and applications of SL, covering the application and hardware layers, the leader election process, and model aggregation techniques. We compare SL with FL and Swarm Intelligence, focusing on data sharing, decentralization, and security. These subsections also explore the role of SL in IoT, improving real-time data processing and security, and conclude with an overview of how SL addresses challenges in distributed learning systems.

\subsection{Concept of SL}

ML can, in theory, be carried out locally with sufficient data and computing resources (Fig.~\ref{fig:learning-models} (A) \cite{warnat2021swarm}). However, data and computation exist separately. In cloud computing, data are centrally transported (Fig.~\ref{fig:learning-models} (B) \cite{warnat2021swarm}), improving the volume of data available for training and improving the outcomes of ML. However, this approach presents challenges such as increased data traffic, duplication, and data privacy concerns. In FL, a central server manages the parameter settings while the data remain at the contributor's location, with local computing performed at the data storage site (Fig.~\ref{fig:learning-models} (C) \cite{warnat2021swarm}). SL eliminates the need for a central server, distributing parameters across the swarm network and developing models locally with private data (Fig.~\ref{fig:learning-models} (D) \cite{warnat2021swarm}).

SL's decentralized structure accelerates training by enabling local data processing at edge nodes, reducing latency, and leveraging computational power across decentralized nodes. It minimizes communication overhead and uses blockchain for secure model updates, while dynamic leader elections optimize training. SL effectively handles non-IID data, improving model robustness and accuracy, and optimizes resource use between nodes with varying computational capacities \cite{warnat2021swarm}.

However, integrating ML into SL can complicate training rate assessments. Variations in ML architecture and complexity affect convergence and efficiency, while SL's decentralized nature and differing computational resources may impact scalability. Blockchain synchronization introduces overhead, and adapting ML methods for SL complicates performance evaluation. Comparative studies with centralized and federated systems are needed to assess SL's real-world benefits.

\begin{figure*}[!t]
    \centering
        \centering
        \includegraphics[width=\linewidth]{Images/The_Comparison_Figure.pdf}
    \caption{Comparative overview of learning models}
    \label{fig:learning-models}
\end{figure*}

\subsection{SL Components}
\begin{figure*}[!t]
\centering
\includegraphics[width=.9\linewidth]{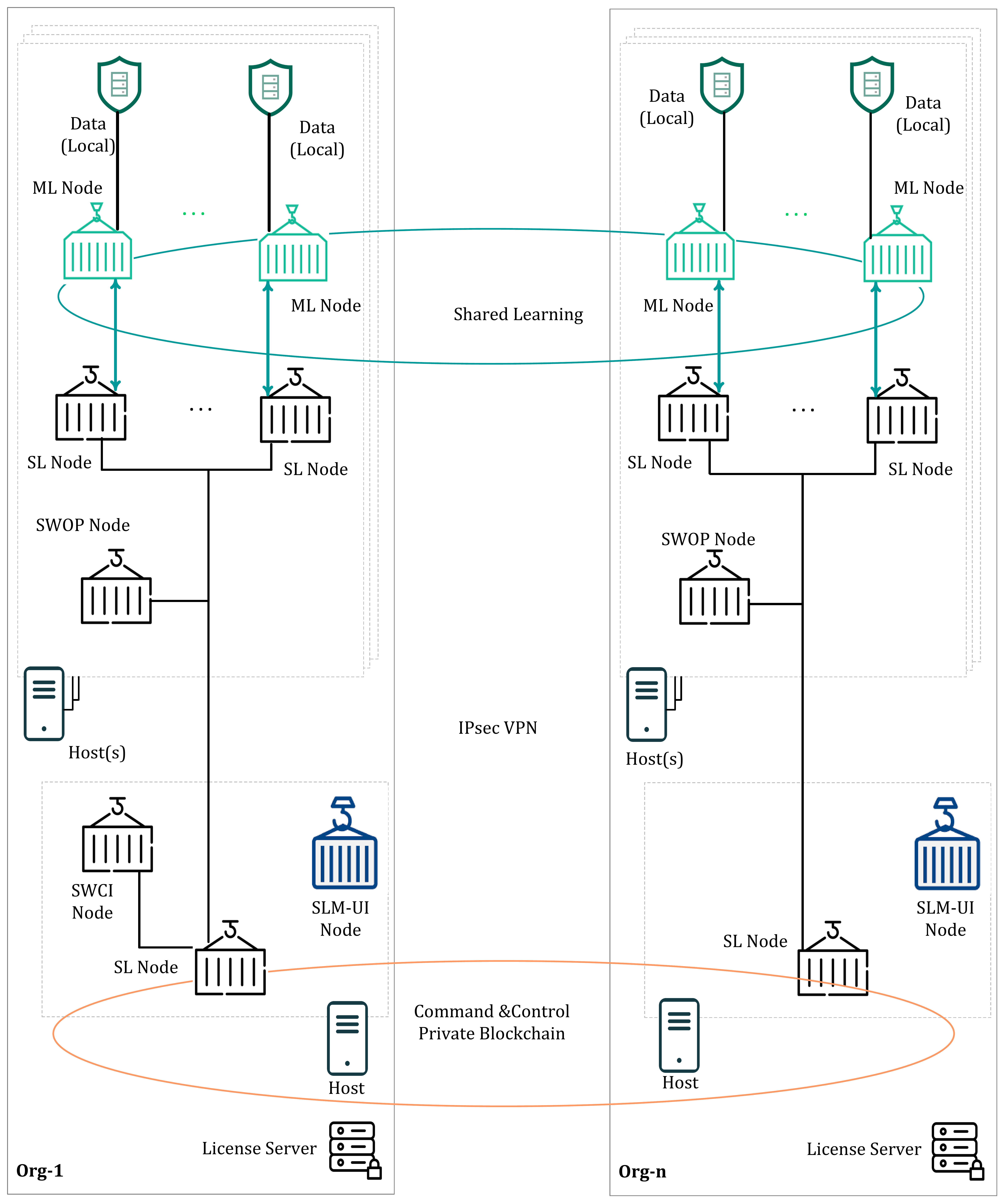} 
\caption{Swarm learning Components}
\label{fig:Swarm learning architecture} 
\end{figure*}

As shown in Fig.~\ref{fig:Swarm learning architecture} \cite{swarmlearning2023}, the SL framework consists of various nodes:
\begin{itemize}
\item \textbf{Swarm Learning (SL) node}: SL nodes run the core of SL, sharing learnings and incorporating insights. 

\item \textbf{Swarm Network (SN) node}: Using the Ethereum blockchain, the SN nodes communicate with each other to track training progress and save global state information about the model. Additionally, during initialization, every SL node is registered with an SN node, and each SN node manages the training pipeline for its corresponding SL nodes. Note that the model parameters are not recorded by the blockchain; instead, it simply stores metadata such as the model state and the training progress.
\item \textbf{Swarm Operator (SWOP) nodes}: SWOP nodes manage SL operations, performing tasks such as starting and stopping Swarm runs, building and upgrading ML containers, and sharing models for training.

\item \textbf{SL Command Interface (SWCI) nodes}: SWCI nodes monitor the framework and can connect to any SN node in a given framework.
\item \textbf{SL Management User Interface (SLM-UI)}: SLM-UI nodes are GUI management tools used to install the framework, deploy Swarm training, monitor progress, and track past runs\cite{swarmlearning2023}.
\item \textbf{SPIFFE SPIRE Server node}: SPIFFE SPIRE Server node ensures the SL framework's security. An SPIRE Agent Workload Attestor plugin is included in each SN or SL node, and it interacts with the SPIRE Server nodes to verify the identities of each node and to obtain and maintain a SPIFFE Verifiable Identity Document (SVID) \cite{han2022demystifying}.
\item \textbf{License Server (LS) node} installs and manages the license to run the SL framework\cite{han2022demystifying}.

\end{itemize}

SL security and digital identity are handled by X.509 certificates, which can be generated by users or standard security software such as SPIRE. SL components communicate using TCP/IP ports, and participating nodes must be able to access each other's ports\cite{swarmlearning2023}.

\subsection{SL Architecture}

SL architecture consists of two primary layers: the application layer, which includes the ML platform, blockchain, and SL Library (SLL), and the infrastructure (hardware) layer, which includes domain-specific data sources and models, such as mission-related or geographic datasets \cite{dovgal2021swarm}. 

The SL system includes two components: Swarm edge nodes and the Swarm blockchain network \cite{chen2021privacy}. Blockchain integration provides several advantages: (1) local data storage, (2) reduced data traffic by avoiding original data exchange, (3) no need for a centralized secure network, (4) enhanced data security against model attacks, and (5) equal rights for all members when merging model parameters \cite{fan2021fairness}. 

\begin{figure}[htbp]
\centering
\includegraphics[width=.8\linewidth]{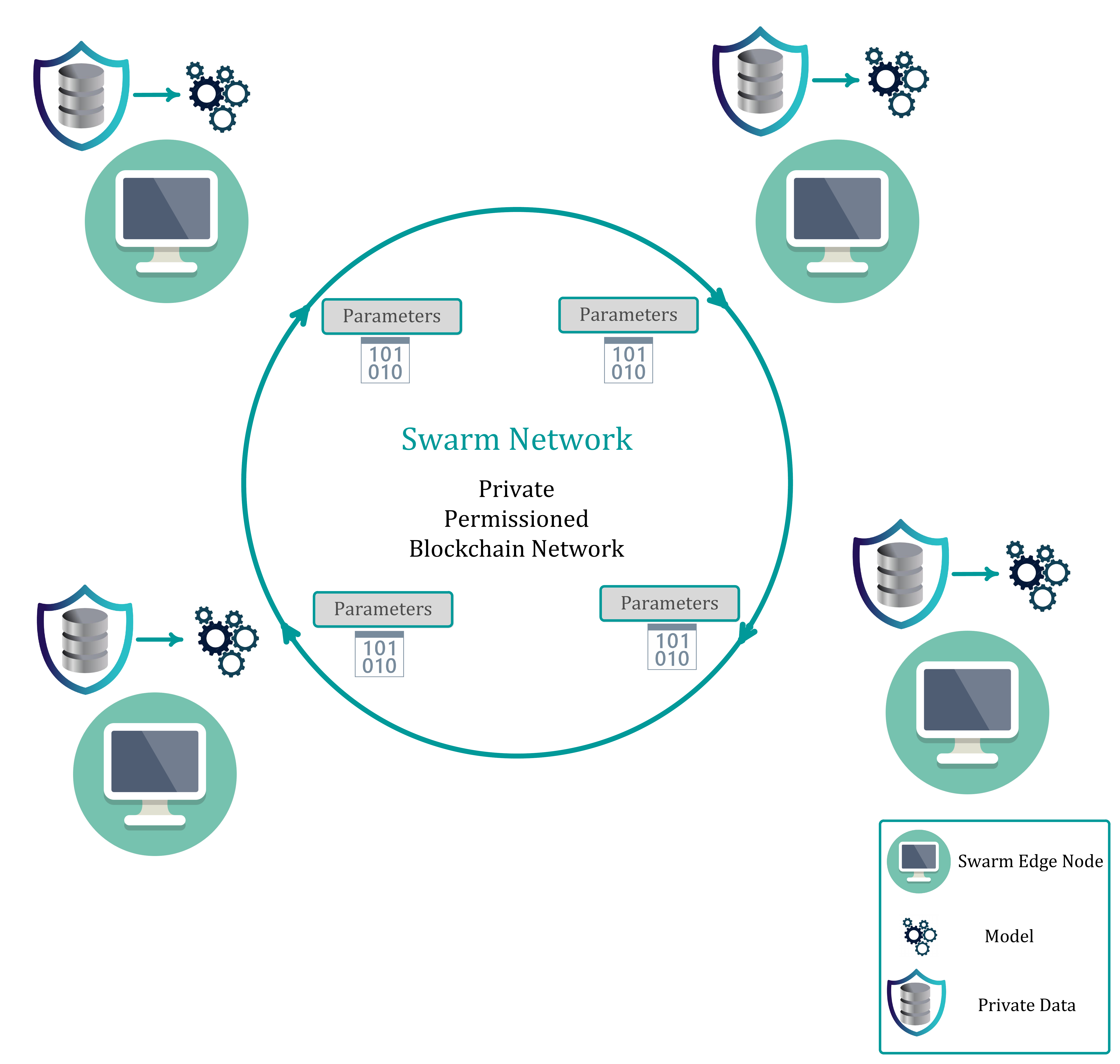} 
\caption{Swarm learning system architecture}
\label{fig:Swarm learning system architecture} 
\end{figure}

Fig. ~\ref{fig:Swarm learning system architecture} \cite{chen2023backdoor, warnat2021swarm, schultze2022swarm} illustrates the architecture of the SL system. The system consists of multiple Swarm edge nodes (Ci), where each node Ci uses local private data Di to train its local model Li. After training, each node shares its model parameters with the network. The nodes are recognized, authorized, and registered with a smart contract in a peer-to-peer blockchain to ensure network security. In each training cycle, selected nodes disclose their model parameters to a Swarm API. A temporary leading node (C) aggregates the parameters into a global model G using a weighted average method \cite{chen2023backdoor}. The process ensures a decentralized and secure collaborative learning mechanism, facilitated by blockchain technology \cite{warnat2021swarm}. 

Swarm edge nodes must first register through the blockchain's smart contract to participate in model training. Once registered, each node downloads the initial global model from the blockchain, trains its local model, and uploads the model parameters to the leader node. The smart contract dynamically selects the leader, which aggregates the parameters of the local model and generates an updated global model, which is then redistributed to the nodes for further training \cite{chen2021privacy}. 

Model sharing in SL is treated as a data transfer process between participating blockchains, which requires secure, consistent, and adaptable methods of interaction. Traditional solutions, such as third-party trust entities or centralized hubs such as Cosmos, are incompatible with the decentralized nature of SL \cite{qi2023game}. 

The workflow for updating the model in SL, shown in Fig. \ref{fig:Workflow of swarm learning} \cite{qi2023game}, consists of two stages. In the first stage, individual organizations update their local models within their own SL nodes, and these updates are stored on their respective permissioned blockchains. In the second stage, multiple blockchains are used to refine local models and synchronize the global model, promoting decentralization and improving security by reducing dependence on external entities \cite{qi2023game}.

\begin{figure}[htbp]
\centering
\includegraphics[width=.75\linewidth]{Images/Workflow_of_swarm_learning.pdf} 
\caption{Workflow of SL with multiple permissioned blockchains. The chains of different colors belong to different participating organizations}
\label{fig:Workflow of swarm learning} 
\end{figure}

\subsection{Leader Election Algorithm (LEA)}
In SL, the fairness and performance of the network are greatly affected by the leader election process. Swarm edge nodes in SL are best placed on instances with plenty of bandwidth and processing power to handle the demands of decentralized decision-making. However, the unfairness of the leader election mechanism could cause nodes to use excessive amounts of bandwidth, which would result in inefficiencies and possibly bottlenecks. Participants may be unhappy with this discrepancy because they believe it is unfair and because nodes with higher data traffic may be more easily targeted by attackers\cite{han2022demystifying} 

The current LEA speculated to be a Proof of Stake (PoS), relies on leadership election on nodes' stakes or account balances. The authors in \cite{han2022demystifying} recommended switching from PoS to a Proof of Work (PoW) model, in which nodes compete to solve cryptographic puzzles and leadership is established by meeting predetermined hash value requirements. By equating the likelihood of becoming a leader based on processing power, this technique seeks to guarantee a more fair distribution of network load among nodes. Future efforts will focus on collaborating with Hewlett Packard Enterprise (HPE) to enhance the fairness and effectiveness of LEA in SL.

\subsection{Features of SL}
SL encompasses several distinct features that strengthen its application in decentralized settings:
\begin{enumerate}[label=\Alph*.]
    \item \textbf{Privacy Preservation:} SL keeps data at each node, minimizing the risk of privacy breaches and confidentiality.
    \item \textbf{Decentralization:} SL reduces the risk of a single point of failure or data monopoly by eliminating the need for central data storage or authority for model aggregation.
    \item \textbf{Continuous Learning:} Models are continuously updated with new data available at each node, adapting to new conditions such as emerging diseases.
    \item \textbf{Data Diversity and Volume:} SL handles larger and more varied data sets from multiple nodes, improving model robustness and generalization.
    \item \textbf{Collaborative Learning:} Nodes collaborate to train a shared model, benefiting from shared insights without actual data transfer, which is crucial for maintaining patient confidentiality.
\end{enumerate}

\subsection{Swarm Learning vs. Federated Learning}

SL and FL are two distributed learning approaches that enable the aggregation of collaborative models between multiple participating nodes \cite{korkmaz2020chain}, \cite{wang2021edge}. Through a series of training rounds, a global model is generated. Importantly, these techniques ensure that participating nodes do not need to share their proprietary datasets, thus maintaining data privacy. However, there are two fundamental differences between SL and FL \cite{fan2021fairness, zhou2024federated}:

\begin{itemize}
   \item \textbf{Information Transmission:} In FL, participating nodes exchange local model parameters and global model updates with a central server. In contrast, SL employs a peer-to-peer network, leveraging blockchain technology and edge computing to allow secure and equitable communication between nodes, without the need for a central server \cite{chen2023backdoor, saldanha2023direct, saldanha2022swarm, yang2022propagable}.
   
   \item \textbf{Central Server Requirement:} FL relies on a central server to aggregate the model parameters from the participating nodes and generate the global model. In contrast, SL operates without a central server. In SL, each participating node has the opportunity to be selected as a temporary server during each training cycle to compile and aggregate model updates \cite{chen2023backdoor}. Using blockchain-based Swarm networks for decentralized and secure parameter exchange, SL eliminates the need for a central server \cite{fan2021fairness}.
\end{itemize}

\subsection{SL vs Distributed FL/Decentralized FL}

Decentralized FL and SL focus on distributed learning by combining edge computing, blockchain, and peer-to-peer networks. With decentralized FL, the need for a central server is eliminated, enabling direct communication between nodes. Blockchain is used to create a more organized system, with a consensus mechanism that helps to update the global model. SL, which is developed by HPE Enterprise, uses blockchain to ensure data integrity, verify nodes, and maintain traceability, while improving data privacy by keeping data local and using cryptographic methods.

In decentralized FL, nodes work together to train a global model without needing a central coordinator, while SL uses a leader election mechanism to collect updates and update the blockchain. Both approaches decentralize ML and keep data localized, but SL also integrates blockchain for enhanced security and dynamic network management. Both methods are ideal for environments where data integrity and auditability are crucial. Future research could focus on comparing these methods in practice and developing hybrid models that combine the strengths of SL and FL.

Beltrán et al. \cite{beltran2023decentralized} compared Decentralized Federated Learning (DFL) and Centralized Federated Learning (CFL), highlighting the benefits of DFL, such as improved fault tolerance and scalability. They analyzed DFL applications in healthcare, Industry 4.0, mobile services, and more. Hallaji et al. \cite{hallaji2024decentralized} discussed the security and privacy of DFL, highlighting its robustness and the need for continuous research to mitigate risks. Zhang et al. \cite{zhang2024decentralized} and Mohammed et al. \cite{mohammed2023energy} reviewed the integration of DFL with blockchain, discussing its operational workflow, challenges such as communication overhead and recommending further research.

Islam et al. \cite{10239404} introduced FLAG (Digital Twin-Based Drone-Assisted Secure Data Aggregation Scheme) for FL in AIoT. FLAG improves AIoT security using drones for data collection and FL training while ensuring privacy through differential privacy. It stores data securely on the blockchain, with a dual validation system to prevent malicious updates. FLAG also uses Digital Twins for efficient training and aggregation, especially in low-connectivity areas. Unlike SL, which uses peer-to-peer collaboration, FLAG enhances security with drone-assisted aggregation and blockchain, ensuring data integrity and privacy with differential privacy and multilayer authentication, making it more suitable for high-threat, low-connectivity environments.

Zhu et al. \cite{10720160} proposed T-BFL, a framework that enhances FL security and efficiency. It integrates a Decentralized Reputation Management (DRM) system for secure model aggregation and optimized blockchain consensus. T-BFL adjusts the difficulty of consensus in the on-chain phase and the reputation of the node in the off-chain phase, while also presenting a dynamic energy allocation algorithm. Although both SL and T-BFL use blockchain for decentralized learning, T-BFL focuses on optimizing consensus and resource allocation, while SL emphasizes secure data management and aggregation. Yu et al. \cite{10795238} examined the privacy benefits of decentralized FL compared to centralized FL, particularly in the context of distributed optimization. Their analysis shows that decentralized FL offers better privacy protection, minimizing information leakage through local gradient exchange and aggregation. Unlike SL, which uses decentralized FL but lacks distributed optimization, this approach includes advanced privacy mechanisms such as gradient differences and privacy limits, emphasizing privacy advantages over SL’s peer-to-peer collaboration model.

Tan et al. \cite{10599518} proposed FU-ISTANet-L, a federated learning model for efficient CSI feedback in distributed edge networks. It minimizes communication overhead by sharing only model parameters and handling heterogeneity of data. Compared to SL, which focuses on decentralized aggregation, FU-ISTANet-L improves efficiency by reducing communication costs, offering advantages in non-IID data and privacy-sensitive communication. Xiao et al. \cite{10463181} introduced FeCGAN, a distributed generative adversarial network for data augmentation in vertical federated learning. The model addresses insufficient data overlap between participants using a central generator and distributed discriminators to generate synthetic data while maintaining privacy. Compared to SL, which focuses on decentralized aggregation, FeCGAN emphasizes data augmentation and heterogeneous data processing, making it more suitable for environments where data alignment between participants is limited.

Wei et al. \cite{wei2025privacy} introduced the differential privacy-based DP-ZOCOA algorithm for unbalanced networks that vary in time in FL. This approach integrates differential privacy, ensuring privacy during federated model training without explicit gradients. Unlike SL, which focuses on decentralized aggregation, DP-ZOCOA emphasizes privacy and optimization in unbalanced, dynamic environments, enhancing efficiency and privacy protection. Liang et al. \cite{ocae313} proposed ODACT, a privacy-preserving algorithm for communication-efficient survival analysis in multicenter studies. ODACT minimizes communication rounds while performing distributed survival analysis, ensuring data privacy. Unlike SL, which focuses on decentralized aggregation, ODACT reduces communication overhead by aggregating data locally and sharing minimal updates, making it more suitable for multicenter healthcare applications.

The choice between DFL and SL depends on specific requirements such as security, scalability, flexibility, data privacy, and cost. SL is ideal for highly regulated fields, such as healthcare and finance, where data integrity is critical. DFL offers better scalability and flexibility for resource-constrained environments but may be more complex to implement. SL provides enhanced security and is suitable for applications that require strict data provenance, while DFL offers more flexibility in handling data privacy and scalability challenges. Table \ref{tab:Comparison of SL and Distributed FL/Decentralized FL} compares SL  with Distributed FL/Decentralized FL, highlighting their key features, privacy mechanisms, blockchain use, and their distinctive advantages in different applications.

\begin{table*}
\centering
\caption{Comparison of SL and Distributed FL/Decentralized FL}
\label{tab:Comparison of SL and Distributed FL/Decentralized FL}
\footnotesize
\begin{tabular}{|p{4cm}|p{5cm}|p{6cm}|}
\hline
\textbf{Feature} & \textbf{SL} & \textbf{DFL} \\ \hline
\textbf{Central Server} & No central server; uses blockchain for communication and decision-making between nodes. & No central server; nodes share model updates, typically via a server for coordination. \\ \hline
\textbf{Use of Blockchain} & Yes, blockchain offers data protection, trust in nodes, and model update security. & Blockchain may be used but is not central to DFL; it may help organize model updates and ensure consistency. \\ \hline
\textbf{Privacy and Security} & High privacy as data stays local; uses cryptography for added security. & Privacy may vary based on setup; some DFL approaches could involve centralized components for aggregation, raising privacy concerns. \\ \hline
\textbf{Data Sharing} & Only model updates are shared, keeping data private and reducing leak risks. & Only model updates (gradients) are shared, but depending on the setup, central server involvement might expose data. \\ \hline
\textbf{Scalability} & Highly scalable, works well in large decentralized networks without a central server. & Scalable, but may face issues with bandwidth and coordination in large networks without blockchain. \\ \hline
\textbf{Leader Election} & Uses a leader election process where nodes decide who will manage model updates. & No designated leader; updates happen between nodes, sometimes with a central server’s involvement. \\ \hline
\textbf{Consensus Mechanism} & Blockchain ensures consensus, preventing fraud and ensuring alignment across nodes. & Consensus mechanisms may be used but are less critical in DFL; it focuses on aggregating model updates. \\ \hline
\textbf{Fault Tolerance} & Very high; if a node fails, others can continue without disruption. & Moderate fault tolerance; depends on the server's capacity, and failure can disrupt the entire system. \\ \hline
\textbf{Best For} & Ideal for privacy-sensitive industries like healthcare and finance. & Suitable for general use cases in industries with fewer privacy concerns, like mobile networks or manufacturing. \\ \hline
\textbf{Implementation Complexity} & More complex; requires blockchain expertise and setting up a decentralized network. & Easier to implement in environments familiar with federated learning and central aggregation. \\ \hline
\end{tabular}

\end{table*}

 \subsection{SL vs swarm intelligence}
Swarm intelligence is a branch of artificial intelligence (AI) that uses the principles of basic agent behavior research to provide algorithms for scheduling, routing, and optimization problems. Particle Swarm Optimization (PSO), Bee Colony Optimization (BCO), and Ant Colony Optimization (ACO) are examples of swarm intelligence algorithms. In contrast, SL is a subset of ML that focuses on distributed and dedicated learning without sharing raw data. SL emphasizes decentralized and collaborative ML in a privacy-preserving manner, while swarm intelligence focuses on problem solving and process optimization, drawing natural influences from a variety of systems.

Although SL and Swarm Intelligence have related names and are inspired by natural swarm behaviors, which may be confusing, it is important to compare them because, in computational and system design settings, they serve distinct purposes and operate on different principles within computational and system design contexts. Exploring the intersections and differences between SL and swarm intelligence can lead to the development of hybrid approaches that leverage the strengths of both. By comparing SL and swarm intelligence, researchers can identify new application areas that may benefit from either approach or a combination of both, aiding in educational and research development. Ultimately, comparing SL and swarm intelligence improves the effective deployment of these technologies in various domains.

SL provides an innovative set of effective solutions to the difficulties of conventional optimization algorithms in swarm intelligence. By addressing these issues, SL overcomes the limitations of traditional optimization algorithms in swarm intelligence and also opens new possibilities to solve complex, dynamic, and large-scale optimization problems in a secure, efficient, and privacy-preserving manner.

The Bacterial Foraging Optimization (BFO) algorithm, introduced by Kevin M. Passino in 2002, is a nature-inspired optimization technique based on E. coli's natural foraging behavior. It has been applied in various fields, including engineering, control systems, and optimization problems. However, BFO has limitations depending on the problem's nature and implementation details, and its performance may not be ideal in all cases. Gan and Xiao \cite{gan2020improved} introduced SL strategies to improve convergence accuracy and prevent premature convergence in BFO. This includes cooperative communication with the best bacteria and competitive learning mechanisms around the world, improving optimal solutions and swarm diversity, and addressing standard BFO deficiencies.

Bolshakov et al. \cite{bolshakov2022deep} have developed a deep reinforcement learning algorithm called Deep Reinforcement Ant Colony Optimization (DRACO), inspired by traditional ant colony optimization and designed for cooperative homogeneous SL. DRACO aims to shape collective behavior in decentralized systems of independent agents, offering an alternative to centralized learning. The advantages of the algorithm include natural parallelization, solving collective tasks beyond the reach of single agents, increased reliability, faster environmental exploration, and economic and energy efficiency.

\subsection{Swarm Learning and IoT}
In traditional cloud-based systems, IoT devices typically send data to central servers to be processed. This results in bottlenecks, privacy violations in transit, and delays. With SL, the processing is local on the device or on nearby edge servers, reducing the quantity of sensitive data that must be sent across the network and shortening the response time. By storing data locally and sharing them securely with the help of blockchain technology, SL improves the privacy and security of data. The method helps to ensure compliance with data protection laws like GDPR by retaining sensitive information in the local sphere. SL also allows IoT devices to learn and adapt in real time perpetually, providing instant analysis and updates. Its decentralized nature also makes it very fault-tolerant, making it an ideal application for healthcare monitoring and industrial automation. It also scales well without the need for a central server, making it suitable for large IoT networks. With the utilization of SL, IoT networks are made more efficient, secure, and complying with privacy regulations, making them ideal for processing large amounts of data. The next section shall elaborate further on the use of SL in IoT applications.

This section outlined SL, emphasizing its decentralized structure and blockchain integration, which enhance security, privacy, and scalability. We compared SL with SL and Swarm Intelligence, highlighting its unique advantages such as serverless architecture and secure collaborative learning. Additionally, SL's potential in IoT was explored, focusing on data privacy and processing speed. In general, SL shows promise as an efficient and privacy-preserving ML solution for industries that require high data integrity and regulatory compliance.

\section{Applications of SL}
\begin{figure*}[!t]
\centering
\includegraphics[width=.8\textwidth]{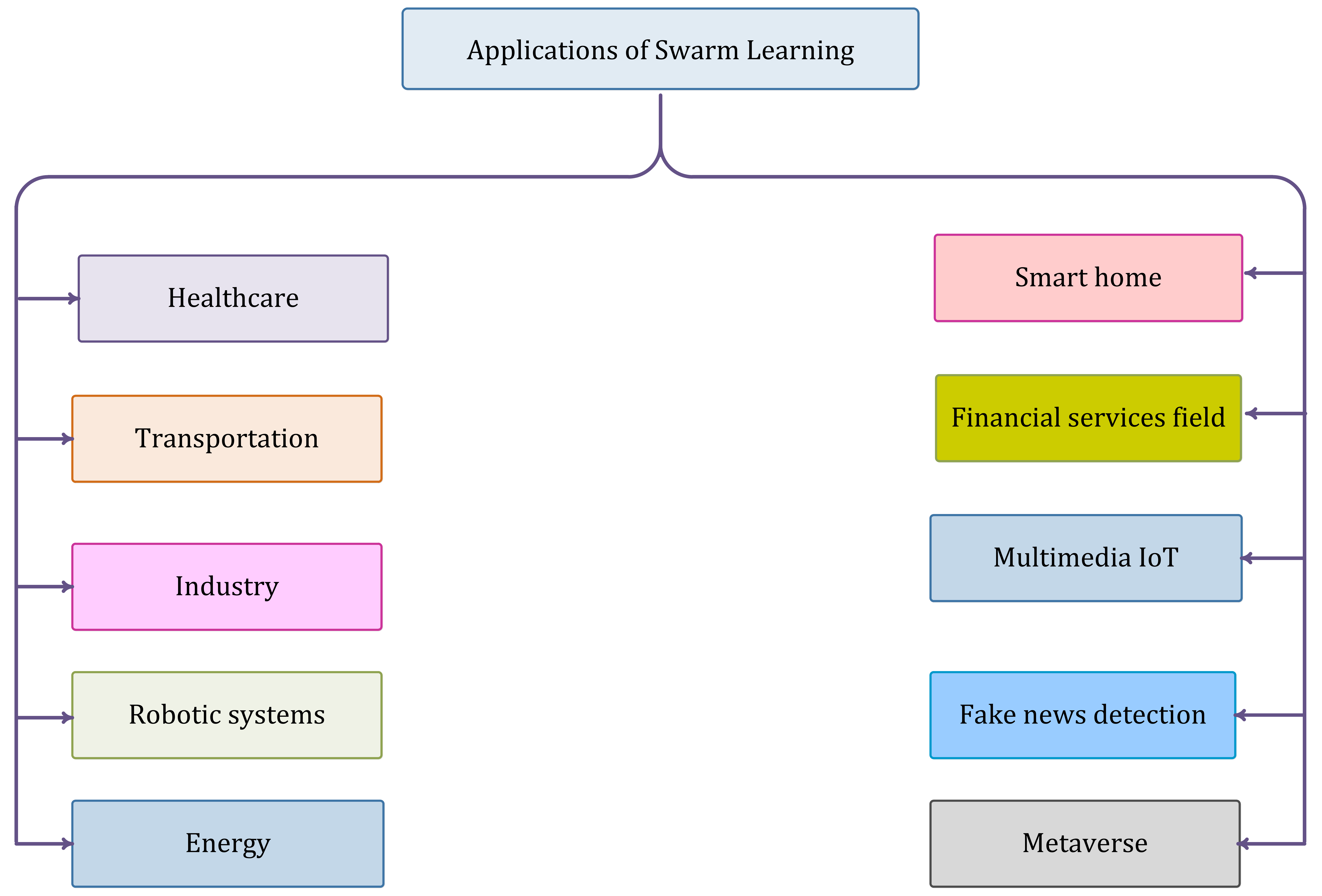} 
\caption{Swarm learning applications}
\label{fig:Applications of swarm learning} 
\end{figure*}

 SL is used in many fields, such as healthcare, autonomous vehicle systems, environmental monitoring, and robotics, as shown in Fig.\ref{fig:Applications of swarm learning}, to improve diagnostic precision, traffic flow, and safety. SL enables data aggregation without compromising privacy, allows communication and learning from experiences, and encourages cooperative robots for complex tasks. Its potential to revolutionize distributed systems and information processing is significant. The following subsections discuss the applications of SL in the reviewed papers.

\subsection{Healthcare}

Modern hospitals generate vast amounts of sensitive patient data, which are restricted by legal and privacy concerns, hindering AI model effectiveness due to small datasets. Distributed DL helps reduce communication and computing costs, but still faces privacy challenges \cite{chen2021privacy}, \cite{becker2022swarm}. SL allows for local ML model training across health nodes, such as hospitals, ensuring data privacy by exchanging and aggregating model parameters without central data collection. The integration of blockchain in SL guarantees data security and confidentiality \cite{becker2022swarm}.

\begin{figure*}[!t]
\centering
\includegraphics[width=\textwidth]{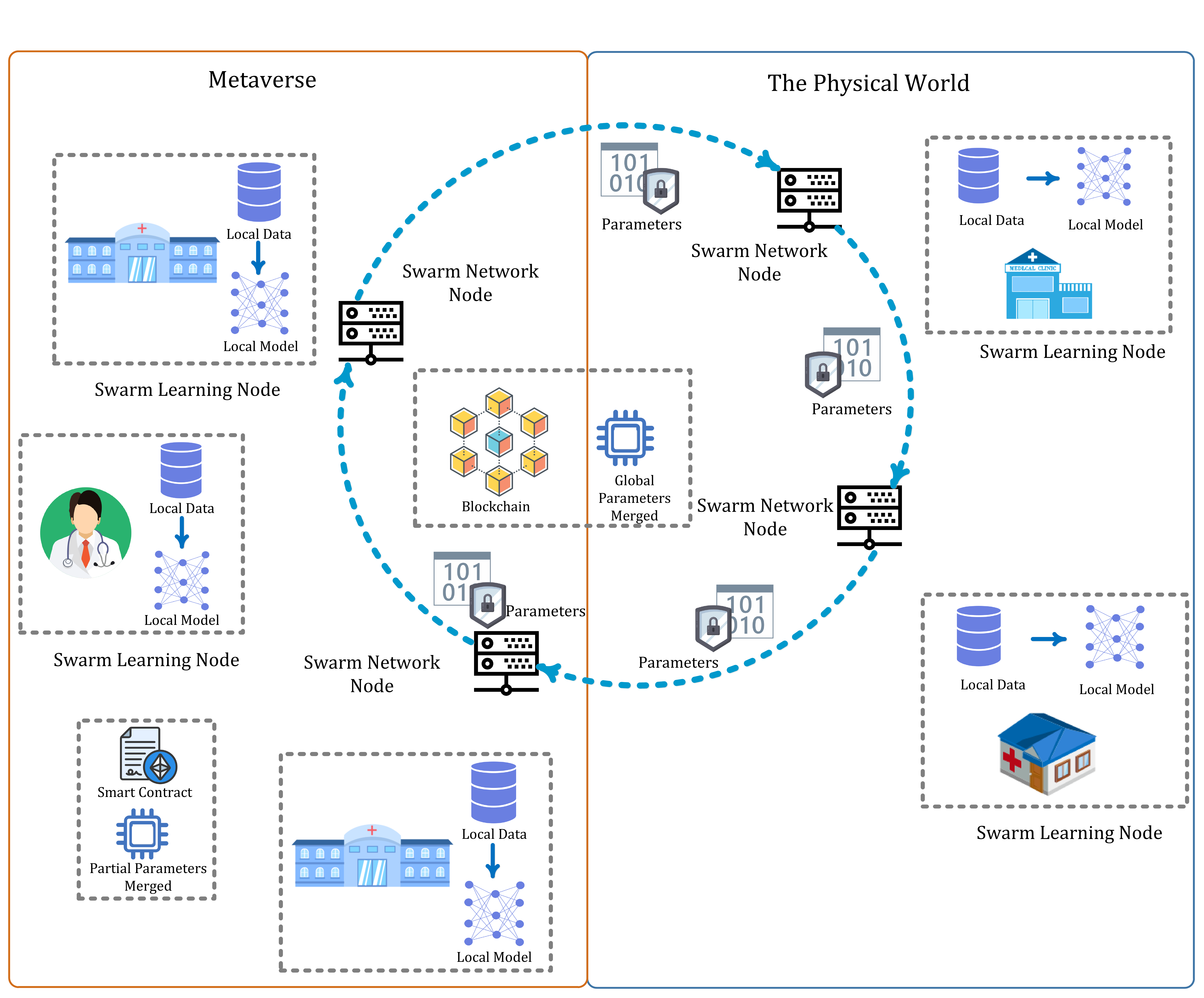} 
\caption{The framework of metaverse swarm learning, which enables cross-domain cooperation between metaverse and the physical world via blockchain}
\label{fig:Swarm Learning in Healthcare} 
\end{figure*}

As shown in Fig.\ref{fig:Swarm Learning in Healthcare}\cite{zhang2023swarm}, the SLN plays a central role by using its unique digital identifier to train local models with private data and contribute to a collective global model. The SNN, pivotal for consensus within the blockchain, manages communication between the SLN and PBN, overseeing the training process, and maintaining the model's status. Lastly, the permissioned blockchain network underpins the model-sharing aspect of SL, safeguarding the security and confidentiality of the process, and facilitating effective collaboration between SLNs.

SL has demonstrated better performance in healthcare applications, such as COVID-19 profiles and chest X-ray images, allowing ongoing learning and enhancement across many data sources while closely respecting privacy laws such as the General Data Protection Regulation (GDPR) and the Health Insurance Portability and Accountability Act (HIPAA). It offers opportunities for the development of cooperative research and diagnostics across hospitals and research institutions networks and is flexible enough to fit a variety of medical data environments. For example, German university hospitals are using SL to evaluate COVID-19 patient data and create AI-based algorithms for the detection of novel biomarkers. SL will become a crucial tool for collaborative healthcare research and precision treatment \cite{chen2021privacy, becker2022swarm, schultze2023building}. 

For example, when hospitals use SL to manage COVID-19 data, they first gather encrypted and anonymized patient data, including symptoms and treatments. Every hospital sets up a separate SL node for safe local data processing. By eliminating raw data exchange, these nodes preserve data privacy by locally training models and sharing just the model parameters over a blockchain. These parameters are then combined by a blockchain consensus method to update and synchronize the global model across all nodes. Real-time deployment of this continually improving model enables more effective diagnosis and treatment plans. SL improves predictive models by integrating diverse datasets from multiple nodes, improving accuracy and treatment efficacy. It prioritizes privacy and security by keeping sensitive patient data on-premises, reducing reliance on central repositories. SL also increases efficiency in hospitals by implementing personalized treatment plans. It is highly scalable, allowing easy integration of new nodes without significant infrastructure changes \cite{warnat2021swarm}. 

Warnat-Herresthal et al. \cite{warnat2021swarm} showcased the effectiveness of Swarm Learning (SL) in training AI models on large histopathology datasets, including over 5,000 patients. They developed disease classifiers for COVID-19, tuberculosis, leukemia, and lung pathologies, demonstrating the potential of SL to enhance medical imaging analysis without need for centralized data collection. Fan et al. \cite{fan2021fairness} explored the fairness issues in SL, specifically within healthcare applications such as the classification of skin lesions. Their work demonstrated that SL outperformed centralized models without amplifying biases, though it also highlighted the complexities of implementing SL. Wang et al. \cite{wang2022generative} introduced SL-GAN, a framework that addresses non-IID data in decentralized ML, advancing the field by enhancing the quality of synthetic data through differential privacy techniques.

Aggarwal et al. \cite{aggarwal2022demed} proposed DeMed, a decentralized privacy preservation system for medical image processing, combining blockchain and self-supervised learning to securely aggregate data, thus addressing challenges in decentralized systems. Chen et al. \cite{chen2021privacy} and Li et al. \cite{li2023privacy} integrated homomorphic encryption with SL, ensuring secure model updates while preserving privacy. Their work introduced novel encryption methods to handle offline participants and mitigate model poisoning risks. Gao et al. \cite{gao2022new} presented a methodology to mitigate the forgetting of global knowledge during local training, improving medical image segmentation with multicenter data. They introduced techniques such as Label Skew-Aware Loss (LaSA) and Feature Skew-Aware Regularization (FeSA) to address label and feature skew.

Yuan et al. \cite{yuan2023cooperative} utilized Swarm Reinforcement Learning (SRL) and blockchain in MEC networks to optimize disease diagnosis, improve task offloading, and partitioning efficiency. Saldanha et al. \cite{saldanha2023direct} applied SL to the identification of gastric cancer biomarkers, suggesting that further research with more biomarkers and larger cohorts would strengthen their findings. Pan et al. \cite{pan2023nanonitrator} proposed an SL-driven system for drug development, improving bioavailability without compromising safety. Saldanha et al. \cite{saldanha2022swarm} also used SL to train models on large histopathological datasets to predict markers of colorectal cancer. Zhang et al. \cite{zhang2023swarm} introduced a framework for sharing secure AI models in metaverse healthcare, addressing security, fairness, and data quality. Mohammed et al. \cite{mohammed2023privacy} developed a decentralized disease diagnosis system using SL for nail images, achieving performance comparable to centralized models.

Shashank et al. \cite{shashank2023swarm} demonstrated the potential of SL as a privacy-preserving approach to analyze breast cancer data from various datasets, improving clinical research while maintaining data privacy. Purkayastha et al. \cite{purkayastha2023general} integrated user feedback into AI model training using SL and a few-shot learning, improving the collaboration between radiographers and AI through continuous training. Shriyan et al. \cite{shriyan2023empirical} employed SL for cataract detection, proposing scalable, privacy-preserving healthcare solutions. He et al. \cite{HE2024171} introduced DP-SL-GAN to tackle non-IID data in decentralized ML by generating synthetic data while maintaining privacy, achieving improved model performance. Finally, Saldanha et al. \cite{Saldanha2025} combined weakly supervised learning with SL for breast cancer detection, showcasing enhanced collaboration between institutions and reducing the need for extensive annotations.

These articles highlight the versatility and transformative potential of SL in diverse healthcare applications, addressing privacy, scalability, and computational challenges, while paving the way for more efficient and secure decentralized machine learning systems. Table ~\ref{tab:Swarm Learning in Healthcare} summarizes the main contributions of these articles.

\begin{table*}
\centering
\caption{Swarm Learning in Healthcare}
\label{tab:Swarm Learning in Healthcare}
\scriptsize
\begin{tabular}{|p{.5cm}|p{1cm}|p{2cm}|p{2.5cm}|p{2cm}|p{2.5cm}|p{3cm}|}
\hline
\textbf{Ref} & \textbf{Application} & \textbf{Contributions} & \textbf{Methodology} & \textbf{Datasets} & \textbf{Key Findings} & \textbf{Future Work} \\
\hline
\cite{warnat2021swarm} & Medical imaging & Demonstrated SL potential for medical imaging analysis. & SL for pathology image analysis & Tuberculosis, leukemia, COVID-19, lung pathologies & Enhanced analysis through multicentric collaboration, maintaining data privacy. & Expand to other medical fields for broader global collaboration. \\
\hline
\cite{fan2021fairness} & Skin disease & Investigated fairness in SL. & SL in skin lesion classification & Skin lesions & Maintained fairness with heterogeneous data without performance loss. & Improve model fairness and address bias. \\
\hline
\cite{wang2022generative} & Clinical settings & Addressed non-IID data challenges in ML. & SL-GAN for non-IID data & Tuberculosis, leukemia, COVID-19 & Overcame challenges in decentralized ML for clinical use. & Further optimize decentralized ML and explore new algorithms. \\
\hline
\cite{aggarwal2022demed} & Medical image analysis & Developed a privacy-preserving decentralized framework. & Blockchain-based SL & Chest X-rays & Used blockchain for privacy-preserving model training. & Expand to more complex tasks, increasing disease variety. \\
\hline
\cite{chen2021privacy} & Privacy-preserving techniques & Applied homomorphic encryption in SL. & Paillier encryption with FedAvg aggregation & MNIST dataset & Advanced secure model updates while preserving data privacy. & Improve defense against model poisoning and handle offline participants. \\
\hline
\cite{gao2022new} & Medical imaging & Solved global knowledge forgetting in local training. & Local knowledge assembly, LaSA loss & FeTS, M\&Ms, MSProsMRI & Addressed Non-IID data issues, enhancing segmentation accuracy. & Apply to systems with unidirectional input constraints. \\
\hline
\cite{yuan2023cooperative} & Disease diagnosis & Cooperative DNN partitioning for disease diagnosis. & Swarm Reinforcement Learning in MEC networks & VGG16, ResNet18 & Accelerated disease diagnosis with minimal latency. & Validate in real clinical settings. \\
\hline
\cite{saldanha2023direct} & Molecular biomarkers & Predict biomarkers in gastric cancer. & SL-trained MSI and EBV models & Bern, Leeds, TUM Cohort, TCGA & Improved biomarker prediction with multicentric data. & Expand to more biomarkers and datasets. \\
\hline
\cite{pan2023nanonitrator} & Drug development & SL-based drug prediction system. & AI-driven discovery & DPN, DDN, DTN & Enhanced therapeutic effects of nitrate. & Clinical trials and pharmacokinetic studies. \\
\hline
\cite{saldanha2022swarm} & Medical imaging & Predict molecular alterations in colorectal cancer. & Retrospective image analysis using SL & Northern Ireland, UK, TCGA & Demonstrated SL feasibility in predicting molecular features. & Expand to other oncology areas and improve scalability. \\
\hline
\cite{zhang2023swarm} & Metaverse healthcare & Secure AI model sharing in the metaverse. & Parameter merging for SL nodes & COVID-19, PAMAP datasets & Enhanced healthcare model reliability with SL. & Improve security and fairness in model-sharing. \\
\hline
\cite{mohammed2023privacy} & Disease diagnosis & Diagnose from nail images using SL. & Transfer learning with SL nodes & Nail disease datasets & High diagnostic accuracy while maintaining privacy. & Expand to other medical data types and refine model training. \\
\hline
\cite{shashank2023swarm} & Cancer research & Utilized large medical data for cancer research. & SL for decentralized cancer diagnosis & WDBC, WPBC, BreakHis & Achieved decentralized learning with privacy adherence. & Expand decentralized training to improve oncology outcomes. \\
\hline
\cite{purkayastha2023general} & Radiology & Enhanced Human-AI partnerships. & SL with user feedback & WDBC, WPBC, BreakHis & Improved personalized radiological assessments. & Optimize Human-AI interaction. \\
\hline
\cite{shriyan2023empirical} & Eye disease detection & Novel cataract detection method using SL. & VGG-19 architecture and SL integration & ODIR dataset & Highlighted SL's advantage over traditional methods. & Expand to detect more diseases. \\
\hline
\cite{HE2024171} & Clinical ML & DP-SL-GAN for non-IID data handling with data augmentation and differential privacy. & SL, differential privacy, teacher-student sharing. & Clinical datasets & Improves performance and privacy over existing methods. & Explore scalability and expand to other domains. \\
\hline

\cite{Saldanha2025} & Medical imaging & Combines weak supervision with SL for breast cancer detection in MRI. & SL for decentralized training & US, Switzerland, UK, Germany, Greece datasets & 3D-ResNet-101 outperforms other models, SL enables multi-center collaboration with data privacy. & Expand SL to other medical fields and improve model robustness. \\
\hline

\end{tabular}
\end{table*}

\subsection{Transportation}
Advancements in communication and computing technologies have greatly improved the Internet of Vehicles (IoV), which plays a key role in improving traffic management, emergency response, and efficiency in Intelligent Transportation Systems (ITS). FL and Federated Deep Learning (FDL) have been introduced to address privacy concerns in IoV applications \cite{hou2023hierarchical}, \cite{wang2023credibility}. However, while SL offers significant advantages, its application in collaborative Vehicle Trajectory Prediction (VTP) presents challenges, including high communication overhead due to global network communication and increased blockchain costs as the network scales, reducing its effectiveness for large-scale deployments \cite{hou2023hierarchical}. 

Lin et al. \cite{lin2022double} proposed a cooperative framework for Vehicle Users (VUs) in the IoV, enabling model training without a central coordinator. Their approach considers VU mobility and incorporates an incentive system based on iterative auctions to encourage participation. The optimization model aims to balance social welfare and market equilibrium, offering a scalable solution for decentralized IoV learning environments. Fan et al. \cite{fan2023cb} introduced CB-DSL, a robust SL framework designed to address data heterogeneity, communication constraints, and security challenges in edge IoT networks. This innovative approach combines biological intelligence with AI to enhance model performance for traffic management, demonstrated through SUMO simulations and real-world healthcare data. CB-DSL improves forecast accuracy and optimizes real-time traffic signal control.

To overcome the limitations of FDL in IoV environments, Wang et al. \cite{wang2023credibility} developed IoV-SFDL, a framework that integrates SL with FDL. The framework addresses communication overhead, vehicle movement, and erratic network conditions by leveraging credibility-based weight adjustment to accelerate model convergence, making it more effective for IoV applications.  Hou et al. \cite{hou2023hierarchical} introduced Hierarchical SL (HierSL), an edge-assisted framework for Vehicle Trajectory Prediction (VTP) in IoV systems. HierSL optimizes collaborative learning by reducing the reliance on global communication and blockchain, offering significant performance improvements over traditional SL and centralized learning, as demonstrated through tests on the NGSIM US-101 dataset. 

Yin et al. \cite{yin2023multi} proposed a Multi-Region Asynchronous SL (MASL) framework that combines blockchain, edge computing, and FL to address the challenges of non-IID data and secure data sharing in large-scale IoV networks. MASL improves anomaly detection and improves data privacy by coordinating intra-regional and cross-regional sharing, making it a valuable solution for IoV systems. To mitigate urban traffic congestion, Liu et al. \cite{liu2023swarm} presented DDaaS, a 6G-enabled solution. The system employs SL for parameter training, congestion prediction, and signal light management, demonstrated through SUMO simulations. DDaaS successfully reduces traffic congestion and improves prediction accuracy, offering a scalable approach to smart city traffic management. 
Mishra et al. \cite{mishra2023swarm} proposed a privacy-preserving SL approach for autonomous driving, addressing concerns related to the use of sensors and cameras. The framework allows secure model training and sharing across nodes, ensuring privacy while maintaining performance, and offering a more effective solution compared to traditional methods and other distributed machine learning approaches. Huang et al. \cite{huang2023dag} created a Directed Acyclic Graph (DAG)-based SL framework for IoV that integrates blockchain and edge computing to ensure secure and asynchronous model training. Their approach addresses challenges such as unreliable communications and malicious attacks by using a dynamic vehicle association algorithm to enhance model training efficiency while ensuring data privacy and security.

Table ~\ref{tab:Swarm Learning in Transportation} shows the main contributions of these articles.

\begin{table*}
\centering
\caption{Swarm Learning in Transportation}
\label{tab:Swarm Learning in Transportation}
\scriptsize
\begin{tabular}{|l|p{1.5cm}|p{2.5cm}|p{3cm}|p{1.5cm}|p{3cm}|p{3cm}|}
\hline
\textbf{Ref} & \textbf{Application} & \textbf{Contributions} & \textbf{Methodology} & \textbf{Datasets} & \textbf{Key Findings} & \textbf{Future Work} \\
\hline
\cite{lin2022double} & IoV & Framework for cooperative model training without central coordinator & Cooperative SL with incentive mechanism & - & Communication-efficient; enhances social welfare & Improve incentive mechanisms for fair participation \\
\hline
\cite{fan2023cb} & Edge IoT & Byzantine-robust Distributed SL for traffic management & Evaluates model performance under i.i.d. and non-i.i.d. conditions & CIFAR-10, MNIST & Improves accuracy, addresses local optima & Validate in real-world settings and tackle more challenges \\
\hline
\cite{wang2023credibility} & IoV & IoV-SFDL: Overcomes communication overhead and privacy issues in IoV & Integrates SL into Federated DL framework & NGSIM & Enhances model convergence speed & Expand to more dynamic IoV scenarios \\
\hline
\cite{hou2023hierarchical} & VTP in IoV & Edge-assisted VTP framework & Hierarchical SL with two-layer learning & NGSIM US-101 & Reduces global communication reliance, improves security & Optimize system topology for better efficiency \\
\hline
\cite{yin2023multi} & IoV & Secure framework for large-scale data sharing in IoVs & MASL with hierarchical blockchain & GTSRB & Improves scalability and security, preserves privacy & Improve training methods, expand blockchain integration \\
\hline
\cite{liu2023swarm} & ITS & DDaaS for traffic congestion reduction & SL-based traffic simulation & - & Improves traffic management, reduces congestion & Enhance traffic control algorithms \\
\hline
\cite{mishra2023swarm} & Autonomous Driving & SL-based training method for autonomous driving systems & SL for privacy and performance & Kitti 3d & Superior privacy and potential performance gains & Compare with other distributed techniques \\
\hline
\cite{huang2023dag} & IoV & Improves data sharing and model training in IoV & Directed Acyclic Graph-based SL with edge computing & GTSRB & Enhances training and attack detection & Develop more robust attack detection mechanisms \\
\hline
\end{tabular}
\end{table*}

\subsection{Industry }
The Industrial Internet of Things (IIoT), driven by IoT, big data, and digital twin (DT) technologies, improves productivity and interoperability in advanced manufacturing. However, DT faces challenges in adapting to dynamic industrial environments and security concerns \cite{xiang2023digital}. SL is transforming manufacturing by enabling intelligent real-time agents for resource allocation, production line optimization, and problem resolution without centralized control. This approach supports Industry 4.0 and smart manufacturing, creating resilient and intelligent factories for the future \cite{wang2024multi}, \cite{sun2023improved}. However, there is limited research on the integration of SL with IIoT, particularly with regard to reliability in complex industrial environments with high temperatures and noise\cite{xiang2023digital}. 

Pongfai et al. \cite{pongfai2020novel} introduced the Dragonfly SL process (D-SLP) algorithm, a novel approach designed for non-linear feedback control systems, which significantly enhances robustness, performance, and stability. Through comprehensive simulations, the algorithm demonstrated superior control capabilities for permanent magnet synchronous motors, although challenges related to unidirectional input constraints and dead zones persist. In a subsequent paper \cite{pongfai2020pid}, the authors leveraged a deterministic Q-SLP algorithm to optimize the parameters of the PID controller, resulting in improved system performance, stability, and convergence time. The approach was validated against conventional methods such as IPSO and WOA, showing superior convergence and performance in the simulations. In their third contribution \cite{pongfai2021optimal}, an adaptive SLP method was proposed for multiple-input / multiple-output (MIMO) systems, allowing dynamic adjustment of PID controller parameters to enhance stability and resilience. This method was successfully validated using a two-wheel inverted pendulum system.

Sun et al. \cite{sun2023data} propose a diagnostic system for the diagnosis of rotating machinery bearing faults, addressing major concerns of data privacy and scarce labeled data. Their new approach leverages convolutional neural networks (CNNs) and adversarial domain networks to facilitate local training without data sharing. Based on this study \cite{sun2023improved}, they introduced an improved approach to the diagnosis of machine component faults with greater diagnosis accuracy and efficiency through the integration of local models such as AlexNet and Chebyshev filters. 

Xiang et al. \cite{xiang2023digital} introduced a paradigm-changing architecture for IIoT systems, with DT technology and credibility-weighted SL being applied to alleviate privacy threats and communication costs. They utilized a DRL approach to optimize system reliability and energy consumption, and an optimization problem was formulated to optimize operational efficiency and sustainability in the IIoT setting. Wang et al. \cite{wang2024multi} proposed an innovative solution to optimize robot assembly cells through cooperative multi-agent SL with DT technology. This model considers each component as an autonomous agent that dynamically responds to issues resulting from mechanical structure, networked software, and hardware integration. The approach supports dynamic reconfiguration, which allows manufacturing systems to respond effectively to dynamic production demands and cycles.

Luo and Zhang \cite{luo2024blockchain} suggested a blockchain-based approach to manage engine data integrity with a focus on preventing tampering and data deletion. Using SL, the solution ensures the privacy of the model parameter while enabling collaborative learning, thus making the best use of the data and improving the credibility of the system. Jiang et al. \cite{10659144} presented a hybrid model driven by SL and blockchain techniques to attain collaborative and secure industrial big data analytics. IFBDAchain framework supports privacy protection and data integrity between various factories based on Hyperledger fabric. It improves fault classification and KPI prediction performance even in non-IID and imbalanced data settings and strengthens collaboration with robust privacy mechanisms and verification methods. Haodong et al. \cite{10852543} proposed a data privacy-preserving diagnostic algorithm for industrial robot harmonic reducers, in which CNNs are integrated with SL. This architecture shares model parameters rather than raw data, improving computational efficiency and fault diagnosis accuracy in industrial environments without a central server, demonstrating its huge potential for practical use.

These SL-based architectures bring considerable improvements to industrial systems, improving efficiency, security, and scalability, and solving issues like data privacy, communication overhead, and reliability, thus making more robust and intelligent operations possible.

\begin{table*} 
\centering
\caption{Swarm Learning in Industry}
\label{tab:Swarm Learning in Industry}
\scriptsize
\begin{tabular}{|l|p{2cm}|p{2cm}|p{4cm}|p{1cm}|p{3cm}|p{2cm}|p{2cm}|} 
\hline
\textbf{Ref} & \textbf{Application} & \textbf{Contributions} & \textbf{Methodology} & \textbf{Datasets} & \textbf{Key Findings} & \textbf{Future Work} \\ 
\hline
\cite{pongfai2020novel} & Nonlinear control systems & D-SLP algorithm for tuning control parameters with secure data sharing & Dragonfly algorithm with SL protocols & - & Superior control performance & Explore specific system constraints. \\
\hline
\cite{pongfai2020pid} & PID controller optimization & Optimized PID autotuning with deterministic Q-SLP algorithm & Swarm and learning for PID parameter refinement & CPC system & Improved convergence and performance & Not specified \\
\hline
\cite{sun2023data} & Diagnostic frameworks & Framework for bearing fault diagnosis in rotating machinery & Adversarial domain networks with CNNs & CRWU, HITsz, XJTU-SY, SCU & Improved fault diagnosis efficiency & Not specified \\
\hline
\cite{xiang2023digital} & IIoT & IIoT architecture enhanced by credibility-weighted SL & Digital Twin and credibility-weighted SL & MNIST dataset & Increased system reliability & Address practical concerns for operational efficiency \\
\hline
\cite{wang2024multi} & Robotic assembly cells & Optimized layout for reconfigurable robotic assembly & Multi-agent SL with digital twin & - & Improved layout optimization & Adapt to rapid manufacturing changes \\
\hline
\cite{luo2024blockchain} & Engine lifecycle data management & Blockchain-based method for data integrity & Blockchain with trusted application for validation & NASA open dataset & Enhanced data integrity & Optimize collaborative learning and data usage \\
\hline
\cite{10659144} & Industrial analytics & Proposes IFBDAchain for secure federated learning with blockchain. & SL with Hyperledger fabric. & Fault classification, KPI prediction & Outperforms local learning in accuracy with non-i.i.d. data. & Improve scalability, real-time applications, and security. \\
\hline
\cite{10852543} & Industrial robotics & CNN-integrated SL for privacy-preserving fault diagnosis. & CNNs with SL for decentralized training & Harmonic reducer dataset & Improved accuracy and efficiency without a central server. & Expand to other systems and optimize integration. \\
\hline
\end{tabular}
\end{table*}

\subsection{Robotic systems}
Collaborative learning among multiple robots can significantly accelerate learning processes by forming a swarm, allowing them to share knowledge centrally or decentrally. In the context of SL, the nodes within the network exchange locally learned models without relying on a central authority. For networked robotic applications, a group of interconnected robots can either collaborate or work independently to complete tasks. Rangu and Nair \cite{rangu2023mobile} proposed a method that uses mobile agents to execute SL on a group of robots, where each agent learns a task. The approach employs a distributed learning model, with the mobile agent compiling and disseminating the learned models locally as it moves through a network of both simulated and real robots. Their experiment demonstrated the effectiveness of SL by using a mixed group of robots, recognizing the prohibitive cost of deploying a swarm of only real robots. By applying reinforcement learning at the local level, SL was proven to be efficient and practical in networks comprising both simulated and real robots.

\subsection{Energy}
The energy sector is adopting ML for forecasting, but privacy and scalability issues hinder centralized approaches. SL, a decentralized blockchain method, enables privacy-preserving collaboration between organizations, improving the accuracy of forecasts without sharing raw data. This section explores the benefits of SL in improving model performance and ensuring data privacy\cite{XU2025125053}. Lei et al. \cite{XU2025125053} proposed a distributed SL strategy for energy sector problems, assuming cooperative, trustworthy parties with high data privacy. The approach addresses photovoltaic power generation and geophysical well-log prediction, analyzing data volume impact, local epoch variation, and security concerns. The results show that SL enables high-performance, secure collaborative learning without a central server, although limitations are discussed.

\subsection{Smart home}
Edge Intelligence (EI) integrates edge computing and AI in smart homes, real-time video analysis, and precision agriculture. However, centralized ML models have limitations such as data privacy breaches and communication overhead\cite{xu2023cooperative}

SL is transforming smart home ecosystems by enabling decentralized decision-making processes. This allows smart devices to communicate and learn from each other's experiences, optimizing energy consumption, security, and automation. Smart thermostats, lighting, and appliances can adjust settings based on occupants' habits, ensuring comfort and energy efficiency. SL also allows security networks to analyze data and adapt without human intervention.

Xu et al. \cite{xu2023cooperative} introduced a new cooperative SL framework to overcome central ML issues by leveraging decentralized SL to predict thermal comfort. This approach reduces communication overhead and improves model performance by leveraging real data from all nodes within the edge computing network. The framework's effectiveness was demonstrated through an extensive empirical investigation using a Non-IID thermal comfort dataset.

Liu et al.\cite{liu2023swarm2} developed ADONIS, a framework to detect abnormal behavior in IoT devices. It uses SL, knowledge distillation, and human-computer interaction (HCI) to improve security and operational efficiency. The decentralized approach reduces the risk of central node failure and reduces latency and energy consumption. ADONIS can be applied to smart cities and IoVs, and its adaptability makes it suitable for various applications. Future research includes further enhancements and refinement of parameter aggregation methods.

\subsection{Financial services field}
By using decentralized networks for data analysis, decision making, and risk management, SL is completely changing the financial services industry. The decentralized nature of SL transforms data-driven decisions in the complex financial landscape. Enhance investment recommendations and fraud detection rates while protecting against single-point failure. Using SL, financial organizations can modify their strategy in response to current market conditions and consumer trends. John et al.\cite{john2023swarm} used SL for credit scoring in Peer-to-Peer lending on a blockchain platform in the financial services industry, ensuring user data privacy and secure transactions. The decentralized model training and credit scoring process eliminate centralized data storage risks. Future work includes testing with real-time datasets and improving user experience.

\subsection{Multimedia Internet of Things}
By enabling the processing and dissemination of decentralized content in real-time in environments containing IoT devices, SL is transforming the Multimedia IoT ecosystem. This method guarantees that the content is personalized for each user, minimizes latency, and maximizes the utilization of network capacity. Furthermore, processing data locally on devices improves security and privacy by reducing the possibility that private information will be hacked.

Zhang et al.\cite{zhang2023improved} have improved the privacy and security of multimedia IoT devices using Radio Frequency Fingerprinting (RFF) for identity authentication. They integrated differential privacy, specifically the Gaussian mechanism, into SL to protect RFF data. They also proposed a novel node evaluation mechanism to prevent malicious nodes from affecting the accuracy and integrity of the model. By guaranteeing the security of the underlying IoT devices through enhanced privacy protection in SL, the research paves the way for safe multimedia services.

\subsection{Fake news detection}
Social media has significantly impacted the distribution of information, but the lack of systematic management has led to the spread of fake news. ML techniques like CNN and recurrent neural networks (RNN)can detect fake news, but centralized detection can violate user privacy. Decentralized methods such as SL offer privacy-preserving learning on local data, reducing hacking risks, and allowing users to maintain confidentiality without sharing data \cite{dong2022integrating}. Dong et al.\cite{dong2022integrating}  developed Human-in-the-loop Based SL (HBSL), a decentralized method to detect fake news. HBSL uses SL and human-in-the-loop (HITL) techniques to detect fake news between nodes, ensuring user privacy. It incorporates user feedback, allowing models to be continuously updated. The method was validated using a benchmark dataset (LIAR), which shows its superiority over existing methods.

\subsection{Metaverse}
The Metaverse faces challenges in reliable extended reality (XR) data transmission due to a lack of incentives and untrust among users. To address these issues, a configurable SL-based safe resource trading mechanism is proposed in \cite{pan2023swarm}. This framework includes subchains for decentralized intelligent reflective surfaces (IRS) resource management and intelligent allocation, a smart contract-enabled scheme, and a decentralized federated learning-driven IRS allocation scheme. Experimental results demonstrate the effectiveness of this configurable SL-based resource trading for reliable XR communication.

Table ~\ref{tab:Swarm Learning Applications} shows the main contributions of those articles.

\begin{table*}
\centering
\caption{Swarm Learning Applications}
\label{tab:Swarm Learning Applications}
\scriptsize
\begin{tabular}{|p{1cm}|l|p{1.5cm}|p{2.5cm}|p{3cm}|p{1cm}|p{2.5cm}|p{2.5cm}|}
\hline
\textbf{Field} & \textbf{Ref} & \textbf{Application} & \textbf{Contributions} & \textbf{Methodology} & \textbf{Datasets} & \textbf{Key Findings} & \textbf{Future Work} \\
\hline
Robotic systems & \cite{rangu2023mobile} & Networked robotic applications & SL with mobile agents for collaborative model training & Robots learn locally, mobile agent aggregates models & - & Efficient SL in robot swarm & Not specified \\
\hline
Energy systems & \cite{XU2025125053} & Decentralized collaborative learning for energy forecasting & SL with blockchain for secure, privacy-preserving model training & Energy forecasting, geophysical well logs & - & Outperforms centralized and federated learning with better privacy & Explore scalability and integration with more diverse energy datasets \\
\hline

Smart home & \cite{xu2023cooperative} & Edge intelligent computing networks & Cooperative SL for thermal comfort prediction & Stochastic gradient descent in a cyclic network & Non-IID thermal comfort dataset & Reduced communication, enhanced privacy & Optimize model performance for real-world data \\
\cline{2-8}
 & \cite{liu2023swarm2} & IoT abnormal behavior detection & ADONIS framework for IoT anomaly detection & SL with knowledge distillation & Traffic dataset & Enhanced IoT security and efficiency & Refine parameter aggregation and communication \\
\hline
Financial services & \cite{john2023swarm} & Credit scoring in P2P lending & Blockchain-based SL for credit scoring & Web 3.0 lending platform for secure transactions & Universal Bank dataset & Ensures privacy and secure transactions & Test dynamic datasets and explore other platforms \\
\hline
Multimedia IoT & \cite{zhang2023improved} & IoT device security using RFF & SL with differential privacy for identity authentication & RFF dataset & - & Enhanced privacy and security in IoT & Extend methodologies to broader IoT applications \\
\hline
Fake news detection & \cite{dong2022integrating} & Decentralized fake news detection & HBSL with user feedback for improved detection & Local learning and collaborative updates & LIAR dataset & Improved fake news detection accuracy & Design models tailored to specific nodes \\
\hline
Metaverse & \cite{pan2023swarm} & 6G-Metaverse XR communication & SL for secure resource trading in 6G-Metaverse & SL and blockchain for decentralized resource management & Custom dataset & Effective in XR communication and resource trading & Investigate SL customization for hardware resource management \\
\hline

\end{tabular}
\end{table*}

\section{Challenges} 
SL enables decentralized collaboration, improving efficiency and privacy in applications such as IoT and healthcare. However, it faces challenges such as security risks, non-IID data, and fairness issues, which limit its wider adoption. The following subsections explore these challenges.

\subsection{Attacks on SL}
\begin{figure*}[!t]
\centering
\includegraphics[width=\textwidth]{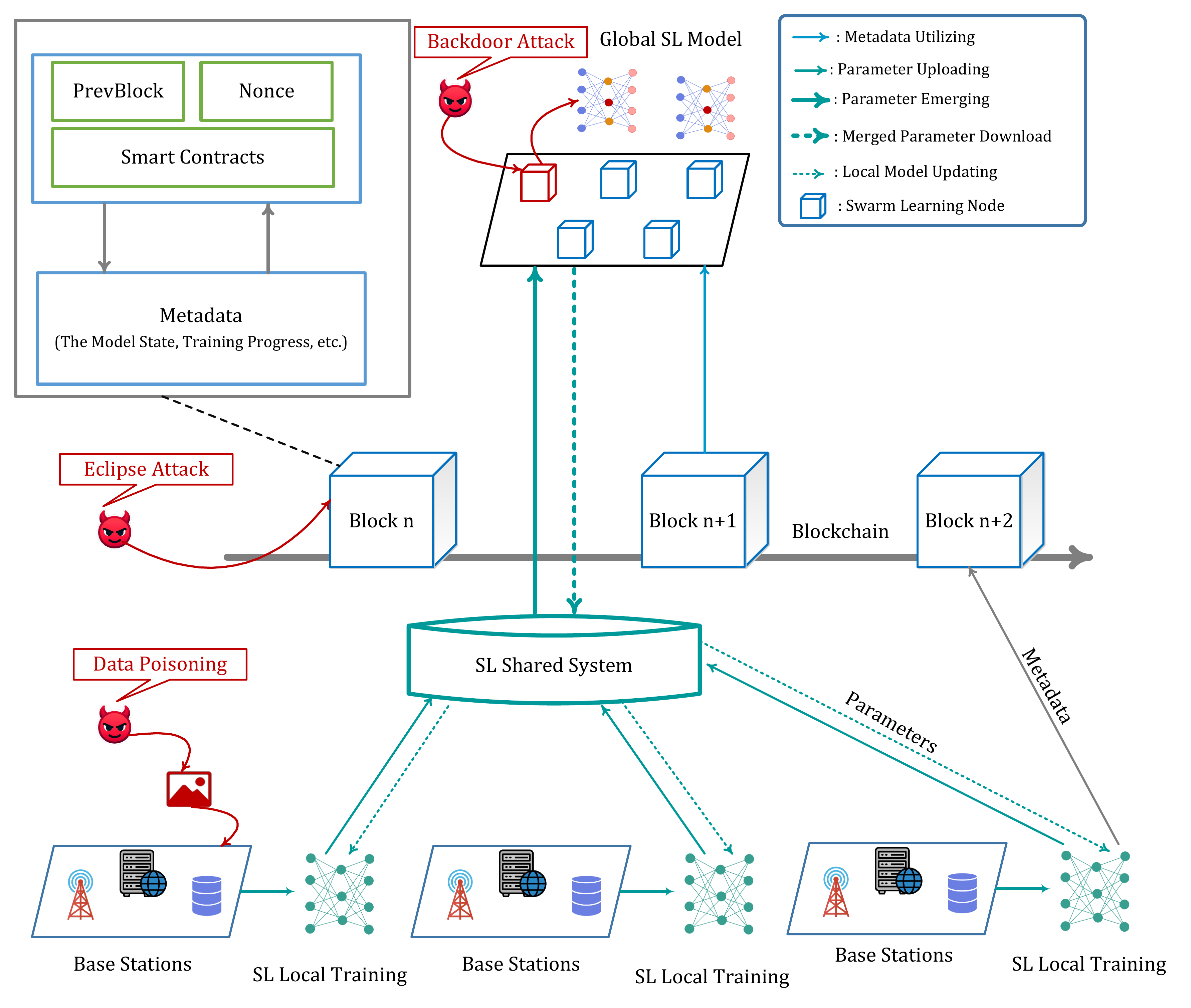} 
\caption{Attack on swarm learning}
\label{fig:Attack on swarm Learning} 
\end{figure*}

SL has the potential to handle distributed large-scale data better than FL, but it also faces significant security issues that require more scrutiny. In the stages of SL, as shown in Fig.\ref{fig:Attack on swarm Learning} \cite{chen2023backdoor}, different attacks can occur: unreliable parties may compromise data during local training and before the locally trained metadata are secured on the blockchain, it might be vulnerable to various network attacks like Eclipse and DDoS. Furthermore, malicious participants could introduce harmful parameters during the merging process, potentially introducing backdoors into the global model. 1) Data poisoning might occur in the local training phase; 2) eclipse attacks could occur in the blockchain P2P network in the metadata upload phase; and 3) the global model could be hacked by poisoned parameters in the parameter aggregation phase \cite{chen2023backdoor}.

\subsubsection{Backdoor attacks against distributed SL}
Despite its advantages in privacy and decentralized training, SL faces significant security challenges, notably backdoor attacks that compromise the integrity and reliability of the system. These attacks manipulate data and training processes, producing incorrect output. The decentralized nature of SL, coupled with non-IID data, complicates the detection of such attacks. Mitigating backdoor attacks requires advanced security measures, robust practices, and innovative collaborative learning techniques to maintain the integrity and trustworthiness of decentralized machine learning environments \cite{chen2023backdoor, yang2022propagable}.

Chen et al. \cite{chen2023backdoor} examined backdoor threats in SL using a pixel pattern backdoor attack method in different datasets (MNIST, CIFAR-10, SVHN). Their study explored various scenarios, including other network sizes, data distributions (IID vs. non-IID), attack targets (single vs. multi-target), and attack types (single-shot vs. multiple-shot). They proposed security strategies such as L2 regularization and noise injection, which were experimentally validated to mitigate the impacts of backdoor attacks.

Yang et al. \cite{yang2022propagable} identified a hybrid vulnerability in SL, where backdoor and eclipse attacks spread backdoors covertly. They introduced the sample-specific eclipse (SSE) strategy, targeting high-contribution nodes to reduce attack costs and accelerate backdoor spread. This study is the first to combine distributed backdoor poisoning with eclipse attacks, revealing how they can work together to stealthily spread backdoors in the SL network. Their proposed strategy focuses on nodes with significant data contributions, enabling faster and more resource-efficient backdoor propagation.

\subsubsection{Poisoning attack} 
SL faces unique challenges from poisoning attacks. Poisoning can compromise the collective learning process, affecting model parameters and performance. The decentralized nature of SL complicates detection, as there is no central authority to monitor data quality or model updates. Therefore, robust decentralized consensus mechanisms are needed to detect and mitigate poisoned inputs \cite{yang2022propagable, zhang2023improved, qi2023game}. Qi Y. et al.\cite{qi2023game} developed strategies to prevent poisoning attacks and ensure the integrity and security of the SL process.

Rongxuan et al.\cite{song2023zero} suggested a Zero Trust Architecture (ZTA)-based defense mechanism for SL to resist poisoning attacks in decentralized learning environments. It identifies a certain flaw in which an attacker 'header' node may poison the model. The defense mechanism is centered around continuous risk calculation and anomaly detection, facilitating dynamic reactions towards threats. The scheme also uses the Manhattan distance and accuracy difference for the detection and avoidance of both header node and edge node threats. The efficiency of the proposed defense method is demonstrated through extensive experiments, demonstrating its implementability in real-world scenarios.

\subsubsection{Eclipse attack}
An Eclipse attack in SL involves an attacker controlling the network communication between nodes. This fits perfectly in peer-to-peer systems where nodes share information and model updates without any central authority \cite{yang2022propagable}. An attacker can isolate a victim node or a group of nodes by controlling their network link, even supplying them with incorrect information or model updates \cite{gupta2024eclipse}. This would impact the integrity of the model and its performance. In order to protect against Eclipse attacks, robust peer discovery and management measures need to be utilized, such as heterogeneous peer connections, peer identity verification, and identifying network patterns that could signify control over communication channels\cite{yang2022propagable}.

\subsubsection{Denial of service attacks (DoDs) attack}
DoS attacks compromise decentralized decision making by poisoning adversarial updates or corrupting the aggregation of model parameters, which can bias decisions in critical applications such as healthcare and autonomous driving. To protect against these attacks, there should be effective consensus mechanisms and anomaly detection techniques to ensure that only reliable updates are incorporated into model aggregation \cite{chen2023backdoor, song2023zero}.

\subsubsection{Sponge attack}. Sponge attack is a DoS attack that takes place in decentralized systems where they are flooded with spurious data, straining system resources and causing significant slowdowns. This interference prevents the ML processes from converging appropriately. Due to the fact that decentralized systems have no centralized point of control, detection and counteractions are difficult, particularly when weaker nodes are attacked. \cite{chen2023backdoor, song2023zero}.

\subsubsection{Inference attacks} 

Inference attacks aim to deduce sensitive information about the training data used by a model, such as recovering private or sensitive attributes. They can be used to determine if a specific data record was part of the training set, infer specific attributes or features of data instances, or attempt to reconstruct a model's parameters. Inference attacks focus on extracting information about the training data or model behavior, such as determining if specific data were used in training or guessing private attributes based on model outputs.  Decentralized ML methods allow multiple nodes to collaboratively learn a shared model without exchanging local data, typically using blockchain technology \cite{rao2024privacy}. Inference attacks exploit shared model updates or the final model to infer properties of the training data or identify unique characteristics of individual participants' datasets. Advanced cryptographic and privacy-preserving techniques such as homomorphic encryption, secure multi-party computation, and differential privacy are employed to protect against inference attacks. However, the balance between privacy protection and model performance is a critical challenge in SL\cite{zhang2023improved}.

\subsubsection{Model inversion attacks}
Model inversion attacks aim directly at reconstructing the inputs used to train the model, effectively reversing the model's computations to approximate or reveal the actual data. They often target models that provide detailed or confident predictions, which can inadvertently reveal information about the training data \cite{fang2024privacy}. While inference attacks often derive indirect information about the data or its attributes, model inversion attacks engage in a more direct and complex effort to recreate the original training inputs themselves. In SL, where nodes collaborate to train a model without sharing their local datasets. The decentralized nature of SL allows each node to contribute to the model's learning by updating it based on local data. However, shared model updates or predictions can leak information, potentially inferring specific characteristics or reconstructing aspects of the original training data. To defend against model inversion attacks, strategies such as output perturbation, differential privacy mechanisms, access controls, and strict query limits can be implemented\cite{zhang2023improved}.

\subsection{Non-IID Problem in SL}
SL enables participants to register, train models, and exchange parameters through edge nodes, ensuring data sovereignty and confidentiality.  However, SL performance is significantly affected by non-independent and identically distributed (Non-IID ) data\cite{wang2022generative}, which can lead to inconsistent model updates and degraded aggregate performance. When data is dispersed unevenly among various network nodes or participants, it is called the non-IID problem in SL. This implies that distinct statistical characteristics, such as mean, variance, and data distribution patterns, may exist in the dataset at each node. Several factors, including variations in patient demographics, the type of medical equipment used, or even the particular focus or specialization of the medical institutes providing the data, could contribute to this heterogeneity in the data. Non-IID data problems include quantity, label, and feature skews. Feature skew and label skew are caused by differences in imaging protocols or demographics, leading to inconsistencies in annotations and Non-IID label distributions. Various strategies, including elastic weight consolidation and batch normalization, have been proposed to address the skews of the features, labels, and quantities in classification tasks. However, these methods do not fully consider label skew, which could cause suboptimal performance\cite{gao2022new}, \cite{liu2023hierarchical}. 

Two types of strategy are now being used to tackle the non-IID challenge: algorithm-based and data-based approaches. Algorithm-based methods align local models with global models, while data-based methods balance distribution but require a trusted central coordinator. Furthermore, with non-IID data, convergence problems may arise when utilizing Generative Adversarial Networks (GAN) for data augmentation\cite{wang2022generative}. 

To address the non-IID problem in SL, methods must be created that can either reduce the impact of data heterogeneity or take advantage of it to increase the global model's resilience and generalizability. Strategies such as advanced aggregation techniques, personalized models, and data augmentation can improve the robustness and generalizability of the global model\cite{gao2022new}. Currently, effective solutions to address the problem of non-IID in SL have not yet been established\cite{wang2022generative}. 

\subsection{Fairness and bias in SL}
Fairness and bias in ML models indicate how they could perform or reflect dominant groupings in the data in an unbalanced way. The impact of SL on model bias and fairness has not yet been fully assessed, even though fairness issues have been considered in the context of FL. In\cite{fan2021fairness}, the authors suggested comparing SL with centralized learning and subgroup-specific model training to investigate the fairness of SL in medical imaging tasks without the need for additional bias mitigation techniques. To provide insight into how SL might balance performance and fairness in healthcare applications, their study seeks to determine if SL's fairness features are more in line with centralized learning or subgroup-specific training.

Future research should focus on developing algorithms that ensure fairness in SL by providing personalized models based on node contributions. This approach needs to balance performance and fairness, encouraging greater participation from nodes with higher contributions. Empirical studies and comparisons with traditional models are necessary to evaluate the effectiveness and scalability of this approach in real world scenarios \cite{TAJABADI2024112451}.

\section{Future Research}
SL addresses privacy and data integration issues, but research gaps exist, indicating potential areas for further exploration.
\begin{itemize}
 \item \textbf{Security and Trust}:  While SL leverages blockchain for security and trust, further research is needed to address potential vulnerabilities, such as advanced cyber threats and insider attacks. Strong trust mechanisms and tailored security measures are crucial for SL networks. Swarm-FHE \cite{madni2023swarm} enhances SL security by integrating fully homomorphic encryption with blockchain, enabling secure collaborative model training even with compromised participants. Additionally, Li et al. \cite{bolshakov2022deep} combine blockchain and lightweight homomorphic encryption to ensure model security, data privacy, and computational efficiency, offering a competitive alternative to Federated Learning in remote ML applications \cite{shang2023decentralized}.
 
 \item \textbf{Dynamic Node Management}: Enhancing the robustness and dependability of SL systems may involve investigating dynamic techniques for node participation and incentive mechanisms to guarantee nodes' continued and productive engagement in the swarm network. 
 \item \textbf{Optimizing Leader Election}: The leader election process in SL can lead to disproportionate bandwidth consumption, inefficiencies, and potential bottlenecks, causing dissatisfaction among participants and potentially compromising network security. To address these challenges, \cite{han2022demystifying}  suggested refining the leader election mechanism for more equitable network load distribution. 
\item \textbf{Scalability and Efficiency}: The ability of SL to expand across a growing number of nodes and a variety of data formats while maintaining efficiency and model performance should be investigated. Enhancing model aggregation techniques and communication protocols could be the main areas of research to facilitate widespread implementations of SL.
\item \textbf{Interoperability and Standards}: For SL to succeed, standards compliance and interoperability amongst various systems are essential. To solve issues with data format, protocols, and compliance, research could examine methods for SL to seamlessly integrate into existing IT systems. Qi et al.\cite{qi2023game} developed a blockchain twin mechanism to improve the interoperability and efficiency of SL on different blockchains, introducing an incentive mechanism for active participation, thus improving the overall performance and security of the SL process.
\item \textbf{Energy Efficiency}: Considering the possible magnitude of SL deployments, especially in the context of IoT, the development of power-saving learning algorithms is of the utmost importance. The emphasis of such research would be on minimizing the energy usage of devices involved in the SL process, a factor that is particularly critical for devices running on batteries or sensors located remotely.
\item \textbf{Cross-domain Applications}: Investigating the potential use of SL in diverse sectors like healthcare, autonomous vehicles, smart cities, and manufacturing can be extremely advantageous. Each of these areas poses distinct challenges and demands, and customized SL approaches could result in significant advancements in how these sectors employ decentralized learning.
\item \textbf{Data Heterogeneity and Non-IID Data}: To efficiently tackle the non-IID issue in SL, forthcoming studies might concentrate on the creation of a hybrid model adaptation method that merges both algorithmic innovations and robust data management strategies. The goal of this method should be to reduce the effects of data heterogeneity and boost the performance and unification of the global model in a distributed environment.
\item \textbf{Advanced-Data Augmentation Techniques}: Investigate the application of advanced generative models, like variational autoencoders (VAEs) or enhanced GANs, to produce synthetic data samples. These samples can efficiently supplement sparse or imbalanced datasets across different nodes, thereby addressing the non-IID problem.
\item \textbf{Ethical AI and Fairness}: As SL models become more widespread, it is crucial to ensure that these models do not perpetuate or exacerbate biases. Research could focus on developing fairness-sensitive algorithms that promote ethical AI practices within SL frameworks.
\item \textbf{Resource Management}: As mentioned in\cite{han2022demystifying}, adding more Swarm coordinator nodes on resource overhead is negligible. However, the resource overhead increases linearly with the number of Swarm edge nodes added, indicating that scaling these nodes should be done carefully. This observation provides valuable guidance and actionable recommendations for developers and researchers applying SL effectively in real-world scenarios. Optimizing the use of computational resources across distributed networks is crucial, especially for resource-constrained environments like IoT devices. More work is needed to reduce communication overhead and improve computational efficiency \cite{10701316}.

\item \textbf{ Deployment Optimization}: Further investigation is needed on how to optimize SL for real-world deployment, particularly concerning the scalability of SL in diverse environments with different network conditions and data distributions \cite{10701316}.

\item \textbf{Integrating ML into SL}: Integrating ML methods into the SL framework can introduce challenges in analyzing the specific contributions of SL to training rate improvements. SL uses blockchain technology to synchronize model updates amongst nodes. Although confidentiality and integrity are guaranteed, the overhead resulting from blockchain operations (such as consensus processes and transaction validations) may outweigh the anticipated gains in training speed from concurrent decentralized training. Therefore, integrating ML methods into SL may complicate the assessment of training rate improvements. Empirical studies and benchmarking against traditional systems are needed to assess its benefits in real-world scenarios.

\item \textbf{Communication overhead}: The Swarm Mutual Learning (SML) framework introduces the Adaptive Mutual Distillation Algorithm and Global Parameter Aggregation Algorithm, which help reduce communication overhead and improve model accuracy \cite{Haiyan2024}. However, optimizing these algorithms for larger datasets and more complex applications remains challenging. Further research is needed to refine the distillation techniques for better knowledge transfer efficiency and explore alternative encryption methods to enhance security while minimizing communication costs. 

\end{itemize}
\section{Conclusion}

SL is a cutting-edge and promising decentralized ML innovation that accommodates secure, efficient, and privacy-preserving collaborative learning independent of data centralization. The review provides invaluable information on the benefits of SL and demonstrates how SL accommodates secure, confidential, and efficient collaborative ML for distributed networks. The main advantages of SL include improved data privacy, reduced chances of centralized data breaches, and the ability to learn from disparate data sources without data transfer, thus avoiding privacy risks. SL can be applied across a wide range of domains, from healthcare to IoV and industrial settings where data confidentiality and learning from distributed and sensitive data are of concern. However, SL must overcome a couple of challenges to reach its full potential.

However, SL must address several challenges to fully realize its potential. Issues such as non-IID data, fairness, bias, and vulnerability to attacks must be carefully considered. To overcome these hurdles, robust decentralized consensus mechanisms and advanced cryptographic techniques are essential to maintain the integrity and privacy of the system. These challenges present significant opportunities for further research and development, offering a broad scope for scholars and practitioners interested in advancing the field of decentralized ML.
\bibliographystyle{unsrt}
\bibliography{references}
\end{document}